\documentclass[10pt,journal,compsoc]{IEEEtran}
\usepackage[ruled]{algorithm2e}
\usepackage{algorithmic}
\usepackage{amsmath}
\usepackage{subfig}
\usepackage{amssymb}
\usepackage{graphicx}
\usepackage{booktabs}
\usepackage{multirow}
\usepackage{multicol}
\usepackage{url}

\ifCLASSOPTIONcompsoc
  \usepackage[nocompress]{cite}
\else
  \usepackage{cite}
\fi
\ifCLASSINFOpdf
\else
\fi
\begin{document}
\title{A fast PC algorithm for high dimensional\\ causal discovery with multi-core PCs}

\author{Thuc~Duy~Le, Tao~Hoang, Jiuyong~Li, Lin~Liu, Huawen~Liu, and Shu~Hu
        % <-this % stops a space
\IEEEcompsocitemizethanks{\IEEEcompsocthanksitem T.D. Le, T. Hoang, J. Li, L. Liu are with the school of Information Technology and Mathematical Sciences, University of South Australia,
Mawson Lakes, SA, 5095.\protect\\
E-mail: Thuc.Le@unisa.edu.au
\IEEEcompsocthanksitem H. Liu is with Department of Computer Science, Zhejiang Normal University, China.
\IEEEcompsocthanksitem S. Hu is Department of Computer Science, USTC, China.}% <-this % stops an unwanted space
\thanks{Manuscript received 2015; revised 2015.}}


\markboth{Journal of \LaTeX\ Class Files,~Vol.~13, No.~9, September~2014}%
{Shell \MakeLowercase{\textit{et al.}}: Bare Demo of IEEEtran.cls for Computer Society Journals}

%\IEEEpubid{\makebox[\columnwidth]{\hfill 0000--0000/00/\$00.00~\copyright~2014 IEEE}%


\IEEEtitleabstractindextext{%
\begin{abstract}
Discovering causal relationships from observational data is a crucial problem and it has applications in many research areas. The PC algorithm is the state-of-the-art constraint based method for causal discovery. However, runtime of the PC algorithm, in the worst-case, is exponential to the number of nodes (variables), and thus it is inefficient when being applied to  high dimensional data, e.g. gene expression datasets. On another note, the advancement of computer hardware in the last decade has resulted in the widespread availability of multi-core personal computers. There is a significant motivation for designing a parallelised PC algorithm that is suitable for personal computers and does not require end users'  parallel computing knowledge beyond their competency in using the PC algorithm. In this paper, we develop parallel-PC, a fast and memory efficient PC algorithm using the parallel computing technique.  We apply our method to a range of synthetic and real-world high dimensional datasets. Experimental results on a dataset from the DREAM 5 challenge  show that the original PC algorithm could not produce any results after running more than 24 hours; meanwhile, our parallel-PC algorithm managed to finish within around 12 hours with a 4-core CPU computer, and  less than 6 hours with a 8-core CPU computer. Furthermore, we integrate parallel-PC into a causal inference method for inferring miRNA-mRNA regulatory relationships. The experimental results show that parallel-PC helps improve both the efficiency and accuracy of the causal inference algorithm.
\end{abstract}

\begin{IEEEkeywords}
Causal discovery, PC algorithm, Parallel computing, High dimensional data, Gene expression data, miRNA targets.
\end{IEEEkeywords}}

\maketitle

\IEEEdisplaynontitleabstractindextext

% \fi
\IEEEpeerreviewmaketitle

\IEEEraisesectionheading{\section{Introduction}\label{sec:introduction}}

\IEEEPARstart{I}nvestigating the associations between variables has long been a main research topic in statistics, data mining and other research areas. For instance, to construct a gene regulatory network, we might want to find the associations between the expression levels of genes. We may use the observed association to explain that one gene regulates the expression of the other. However, the association does not tell us which gene is the regulator and which gene is the target,  or indicate the chance of a third gene regulating the two genes. Hence,  associations between genes may not imply the cause-effect nature of gene regulatory relationships.

\

There has been a long controversy  over whether causality can be discovered from observational data. Many scientists have attempted to design algorithms of inferring causality from observational data over the last few decades \cite{Pearl2000, Spirtes2000,pearl1988, Neapolitan2004}. As a result,  there has been more and more evidence demonstrating the possibility of discovering causal relationships from non-experimental data in different research areas \cite{granger1969,sims1972, Pearl2000}. Especially,  two Nobel prizes were awarded for such causality methods in the field of Economics in 2003 and 2011 respectively \cite{granger1969, sims1972}.  These works have strongly motivated the utilisation of observational data in exploring causal relationships in other important research areas, such as finding the genetic causes of cancers.

\

The standard method for discovering causality is a randomised controlled experiment. For example, to assess the effect of  gene $A$ on other genes, biologists use the gene knockdown experiment. In the experiment, they create two groups of samples, control and transfected groups. Gene $A$ is knocked down in the control group whilst remains in the transfected group. The changes in the expression levels of the target genes between the two groups are the effects that gene $A$ has on them. However, conducting such experiments would incur humongous costs given  that thousands of genes are required to be tested. Moreover  a randomised controlled experiment is generally impossible to conduct or restricted by ethical concerns in many cases. Therefore, discovering causality from observational data is a crucial problem.

\
%There has been a long controversy  over whether causality can be discovered from observational data. Many scientists have attempted to design algorithms of inferring causality from observational data over the last two decades \cite{Pearl2000, Spirtes2000,pearl1988, Neapolitan2004}. As a result,  there has been more and more evidence demonstrating the possibility of discovering causal relationships from non-experimental data in different research areas \cite{granger1969,sims1972, Pearl2000}. Especially,  two Nobel prizes were awarded for such causality methods in the field of Economics in 2003 and 2011 respectively \cite{granger1969, sims1972}. In Computer Science area, the ACM Turing Award was presented to Professor Judea Pearl  in 2011 for his work of building the theory of discovering causality from observed data \cite{Pearl2000}. These works have strongly motivated the utilisation of observational data to explore causal relationships in other important research areas, such as finding the genetic causes of cancers.
%It was shown that causal relationships can be discovered from observational data in many practical cases \cite{Pearl2000causality,Spirtes2000}.

One of the most advanced theories with widespread recognition in discovering causality is the Causal Bayesian Network (CBN).  In this framework, causal relationships are represented with a Directed Acyclic Graph (DAG) \cite{Pearl2000}. Learning a DAG from data is highly challenging and complex as the number of possible DAGs is super-exponential to the number of nodes \cite{Robinson1971}. There are two main approaches for learning a DAG: the search and score approach, and the constraint based approach. While the search and score approach raises an NP-hard problem, the complexity of the constraint based approach is  exponential to the number of nodes. The high computational complexity has hindered the applications of causal discovery approaches to high dimensional datasets, especially gene expression datasets of which the number of genes (variables) is large and the number of samples is normally small.

\

A well-known  constraint based algorithm  is the Inductive Causation (IC) algorithm proposed in \cite{Verma1990}. It is a conceptual algorithm and was implemented in the PC (named after its authors, Peter and Clark) algorithm \cite{Spirtes2000}. The PC algorithm starts with a complete, undirected graph and deletes recursively edges based on conditional independence decisions. For example, the edge between $A$ and $B$ is removed if we can find a set $S$ that does not include $A$ and $B$, and when conditioning on $S$, $A$ and $B$ are independent. The PC algorithm has been implemented in various open-source sofware such as TETRAD \cite{Spirtes2000}, pcalg \cite{pcalg}, and bnlearn \cite{bnlearn}, and has become a reliable tool for causal explorations.

\

There have been several applications of the PC algorithm in Bioinformatics \cite{ZhangX2012, Maathuis2010,le2013, Zhang2014a, Zhang2014b}. Specifically, Zhang et al. \cite{ZhangX2012} applied the PC algorithm which used mutual information for the conditional independence tests to learn the gene regulatory networks from gene expression data. Maathuis et al. \cite{Maathuis2010} used the PC algorithm to learn the causal structure of the gene regulatory networks, then they applied $do$-$calculus$ \cite{Pearl2000} to the learnt network to infer the causal effect that a gene has on the other. Le et al. \cite{le2013} applied the approach in \cite{Maathuis2010} to predict the targets of microRNAs (miRNAs, an important class of gene regulators at the post-transcriptional level). In a similar fashion, Zhang et al. \cite{Zhang2014a} inferred the miRNA-mRNA interactions that had different causal effects between biological conditions, e.g. normal and cancer. These applications of the PC algorithm have been shown to outperform other computational methods that are based on correlation or regression analysis. Moreover, the PC algorithm is able to distinguish direct interactions from indirect interactions in gene regulatory networks \cite{ZhangX2012, Zhang2014b}.

\

However, there are two main limitations of the PC algorithm, especially when applying to high dimensional biological datasets: (i) the runtime of the PC algorithm, in the worst case, is exponential to the number of nodes (variables), and thus it is inefficient when applying to high dimensional datasets such as gene expression datasets, and (ii) the result from the PC algorithm is variable ordered-dependent, i.e. when we change the order of the variables in the input dataset, the result may change.

\

Several methods \cite{silverstein2000,tsamardinos2003,li2013,li2015Tist}  have been proposed to improve the efficiency of the PC algorithm. They aim to introduce efficiency-improved alternative methods, but they rather compromise the accuracy.  These methods either search for only specific causal structures or use heuristic functions to improve the efficiency of the algorithm. However, the results from these heuristic methods are either incomplete  or involving a high rate of false discoveries.

\

Additionally, the order-dependence of the PC algorithm is problematic in high dimensional datasets. Colombo et al. \cite{colombo2012} has shown experimentally that around 40\% of the edges (2000 edges) learnt from a real gene expression dataset are not stable, i.e. these edges only appear in less than half of the results obtained with all the different orderings of nodes. This problem makes the knowledge inferred by using the PC algorithm less reliable. To overcome the order-dependence of the PC algorithm, researchers proposed several modifications to the procedure of the algorithm  \cite{Spirtes2000,colombo2012,dash1999hybrid,cano2008score}. For instance, Colombo et al. \cite{colombo2012} proposed a modified version of the PC algorithm called stable-PC. The algorithm aims to  query all the neighbours of each node and fix these neighbours while conducting conditional independence tests (CI tests henceforward)  at each level (based on size of the conditioning sets) of the PC algorithm. However, this modification requires more CI tests and therefore results in even longer running time for this already inefficient algorithm. Our experiment on a dataset from the DREAM 5 challenge (http://dreamchallenges.org/)  with 1643 variables and 805 samples shows that the order-independence version of the PC algorithm \cite{colombo2012} takes more than 24 hours to run (unfinished). The inefficiency of the algorithm hinders its application in practice, as we would not estimate the runtime of a real world dataset up-front, and thus, prolonged running time would prevent us from applying the algorithm.

\

In another direction, parallel computing provides a straightforward approach to improving the efficiency of the algorithm, but it has been impractical for normal end users. The last decade has  seen a fast growth of parallel computing frameworks, such as MapReduce framework \cite{dean2008mapreduce}. However, such frameworks need suitable cluster facilities and require some solid technical understanding from users. Meanwhile, end users, e.g. biologists, normally use personal computers or small lab servers for their everyday research, and may not easily acquire the required technical knowledge for applying parallelised data mining algorithms. On the other hand, the fast development of computer hardware in the last decade has resulted in the widespread availability of multi-core personal computers. It is common for a modern computer to have a four-core CPU, and small servers to have eight-core CPUs. Therefore, it is of great interest to design parallel algorithms that are suitable for personal computers. Specifically, we aim to design the parallelised PC algorithm (called parallel-PC hereafter) for multi-core personal computers.
%Such algorithms should not require extra knowledge or installation effort from end users.

\
%The limitations of the PC algorithms have also prevented the applications of other causal discovery and causal inference methods too. Most of the methods in constraint based approach are based on the idea of the PC algorithm directly or indirectly. The FCI \cite{Spirtes2000} algorithm and RFCI \cite{colombo2012learning} are based on PC algorithm to learn causal graphs which allows latent variables. CCD algorithm \cite{richardson1996discovery} for learning Markov equivalent class and IDA algorithm \cite{Maathuis2009} for estimating causal effects both utilise the PC algorithm as the first step. Therefore, the family of those algorithms suffer from the same problems of the PC algorithm.
%which results in a practical algorithm for high dimensional datasets like gene expression datasets.These algorithms such as FCI, RFCI, CDC, IDA  are based on the PC algorithm and therefore can be benefited from the proposed algorithm. Moreover, the efficient PC algorithm unblocks the application of a family of causal inference algorithms in high dimensional datasets.

In this paper, we present a method to parallelise the order-independence PC algorithm in \cite{colombo2012} for high dimensional datasets. We propose to parallelise the CI tests at each level  of the algorithm.  The CI tests at each level are grouped and distributed over different cores of the computer, and the results are integrated at the end of each level. Consequently, the runtime of our parallel-PC algorithm is much shorter than the original PC algorithm and the results of the parallel-PC algirhtm is independent of the variable ordering in a dataset. Importantly, it does not require any extra knowledge or installation effort from users. Our experiment results on both synthetic and  real world datasets show that it is now practical to explore causal relationships with the PC algorithm in gene expression datasets using a multi-core personal computer. The proposed algorithm contributes to bridging the gap between computer science theory and its practicality in important scientific research areas. As an application, we modified the causal inference method in \cite{le2013}, by replacing the PC algorithm with the parallel-PC algorithm, to infer miRNA-mRNA regulatory relationships in three different cancer datasets. The experimental results show that our method not only outperforms the original causal inference method that uses the original PC algorithm, but also faster.

\

We summarise the contributions of the paper in the following:
\begin{enumerate}
\item Presenting a fast PC algorithm  which is suitable for high dimensional datasets, and the results are independent of the order of the variables. To the best of our knowledge, this is the first work of utilising parallel computing for the PC algorithm.
\item Providing the software tool and proving that it is reliable and efficient using a wide range of real world gene expression datasets.
\item Modifying a causal inference method to infer miRNA-mRNA regulatory relationships with better performance.
\end{enumerate}

\

The rest of the paper is organised as follows. Section 2 discusses related work, and Section 3 presents the PC algorithm and the stable-PC algorithm as well as the demonstrating examples. The proposed algorithm is presented in Section 4. Experiment results are shown in Section 5, Section 6 presents an application of parallel-PC in inferring miRNA-mRNA regulatory relationships, and finally Section 7 concludes the paper.
\section{Related work}

The well-recognised methods for causality discovery are based on probabilistic graphical  modeling \cite{Heckerman1995,Pearl2000}. The structure of these graphical models is a DAG (directed acyclic graph), with its nodes representing random variables  and edges representing dependencies between the variables \cite{edwards2000}. There are two main approaches to learning the causal structure from data: 1) search and score and 2) constraint based methods. The search and score methods \cite{cooper1992,Heckerman1995,glymour1999} search for all possible DAGs whilst using a scoring function to measure the fit of each DAG to the data. The DAG that best fits the data will be chosen. However, learning Bayesian networks using this approach is an NP-hard problem \cite{chickering1994}, and the proposed methods following this approach were only able to cater for datasets with a limited number of variables, and therefore may not be suitable for high dimensional datasets. Meanwhile, the constraint based approach \cite{Pearl2000,cooper1997,silverstein2000} uses CI tests to remove non-causal relationships between variables. These methods are suitable for sparse datasets in practice \cite{Kalisch2007}.

%In an attempt to increase the scalability of the search and score methods, Friedman et al. \cite{friedman1999b} proposed the heuristic techniques that  iteratively restricts the parents of each variable to being part of a small subset of candidates. Then, it searches for a network that satisfies the constraints. This heuristic method is more efficient than the original
\

Due to the significance of the PC algorithm in the constraint based causal discovery approach, there have been several methods aiming to modify the procedure of the PC algorithm directly to improve the efficiency and/or the accuracy of the algorithm. Steck and Tresp \cite{steck1999bayesian} used the necessary path condition to reduce the number of CI tests in the PC algorithm. They have proved that when testing the edge between $X$ to $Y$, we do not need to condition on the nodes that are not in a path from $X$ to $Y$. This modification may help reduce the chance for making errors by conducting fewer tests. However, it requires more running time for finding the nodes that are not in the path. Other researchers \cite{cooper1997,Abellánsomevariations} proposed to replace the CI test, which is normally based on the Chi-square statistical test, with the Bayesian statistical tests. They aim to reduce the error rates made by the CI tests. However, there is still no clear evidence in real world datasets about the impacts of this replacement on the efficiency improvement of the PC algorithm.

\

Meanwhile, the other constraint based methods \cite{Pearl2000} infer causal relationships by performing CI tests and searching for specific causal structures. For instance, Cooper \cite{cooper1997} proposed the LCD method to learn only a small portion of edges rather than the complete network. Specifically, the LCD algorithm searches for the CCC structures which consist of three pairs of correlated variables: ($A$, $B$), ($A$, $C$) and ($B$, $C$). If $A$ and $C$ become independent when conditioned on $B$, we may infer that one of the following causal relations exists between $A$, $B$, and $C$: $A \leftarrow B \rightarrow C$; $A \rightarrow B \rightarrow C$; and $A \leftarrow B \leftarrow C$. Silverstein et al. \cite{silverstein2000} proposed the similar method that focuses on the CCU structures. The CCU method searches for two correlated pairs $(A, B)$ and $(A, C)$ and an independent pair $(B, C)$. When conditioning on $A$, if $(B, C)$ becomes dependent then the causal relationships $B\rightarrow A \leftarrow C$ are concluded. Compared to the PC algorithm, such specific structure finding methods are more efficient, but the causal discoveries are incomplete.

\

Due to the inefficiency of algorithms concerning causal relationships between all pairs of variables in a dataset, several methods have been proposed to discover only the local causal relationships from data, i.e. the relationships between a specific target and its neighbours. A popular approach is to identify the Markov blanket of a target variable. In a causal graph, Markov blanket has two equivalent definitions: 1) the Markov blanket of a node $T$ is a set $MB(T)$ such that $T$ is independent of other nodes when conditioning on $MB(T)$, or 2) the Markov blanket of a node $T$ comprises its parents, children, and children's parents (spouses). Based on these two definitions, two main approaches have been proposed to discover the Markov blanket, the grow-and-shrink approach and the divide-and-conquer approach. The grow-and-shrink approach is based on the first definition and it uses the heuristic function in the ``grow" step to find the superset of the true Markov blanket, then the false positives in this super set will be removed in the ``shrink" step. Some well-known methods in this approach include GS (Grow and Shrink) \cite{Margaritis99a}, IAMB (Incremental Association Markov Blanket) \cite{tsamardinos2003}, and IAMB variants \cite{yaramakala2005}. The divide and conquer approach uses the topology constraints in the second definition of Markov blanket. This approach firstly finds the parents and children of the target node, then the spouses can be found by searching through the parents and children of the identified nodes in the first step. This approach attracts several methods namely HITON-PC/MB \cite{aliferis2010a}, MMPC-MB \cite{Tsamardinos2003MMMB}, semi-Interleaved HITON-PC \cite{aliferis2010a}, PCMB \cite{pena2007towards}, and IPC-MB \cite{fu2008} (see \cite{aliferis2010a} for a review).

\

In another direction, Jin et al. \cite{jin2012} proposed an efficient method for testing the persistent associations between the target variable and its neighbours. The method is based on partial association tests and association rule mining framework to remove the spurious associations.  In a similar fashion, Li et al. \cite{li2013,li2015Tist} integrated the ideas of retrospective cohort study with the association rule mining framework. They proposed to divide the samples into two groups of individuals, who share common characteristics but differ in regard to a certain factor of interest. The newly-designed dataset is called ``fair dataset" and is used to infer the level of influence that the factor of interest affects the target variable. These works were extended  in \cite{li2015}, and the software tools were also provided. However, these methods are only applicable for binary datasets with a fixed target variable, and thus may not be suitable for high dimensional datasets, e.g. gene expression data.

\

The common characteristic of most of the above-mentioned works is that they aim to propose efficiency-improved alternative methods  to the PC algorithm. However, the results from these heuristic methods may involve a high rate of false discoveries.

\

Scutari \cite{Scutari} utilised parallel computing technique for constraint based local causal discovery methods. Meanwhile, Chen et al. \cite{Chen} proposed a method to parallelise the  Bayesian network structure learning algorithm using search and score approach. However, there is still no existing work that parallelises the PC algorithm.

\section{The original-PC and stable-PC algorithms}
\subsection{Notation, definitions, and assumptions}
%In $G$,  $X_j$ is a \textit{parent} of  $X_i$ if there is a directed edge $X_j \rightarrow X_i$. We use $pa_i(G)$ to represent the set of all parents of $X_i$.
Let $G=(\mathbf{V},\mathbf{E})$ be a graph consisting of a set of vertices $\mathbf{V}$ and a set of edges $\mathbf{E}\subseteq \mathbf{V} \times \mathbf{V}$. The set of  vertices that are adjacent to $A$ in graph $G$ is defined as: $adj(A,G)=\{B: (A,B) \in \mathbf{E}\ or (B,A) \in \mathbf{E}\}$. $B$ is called a \textit{collider} if we have the \textit{$v$-structure}: $A\rightarrow B \leftarrow C$.

\

%A $v$-$structure$ is an ordered triple of vertices, $(X_i, X_j, X_k)$, such that in $G$ there exist  directed edges $X_i \rightarrow X_j$ and $X_j \leftarrow X_k$, and $X_i$ and $X_k$ are not adjacent.  $X_j$ is then known as a \textit{collider} in this $v$-structure.
Graph $G$ is a Directed Acyclic Graph (DAG) if $G$ contains only directed edges and has no directed cycles. The \textit{skeleton} of a DAG $G$ is the undirected graph obtained from $G$ by ignoring the direction of the edges. An \textit{equivalence class} of DAGs is the set of DAGs which have the same skeleton and the same $v$-structures. An equivalence class of DAGs can be uniquely described by a \textit{completed partially directed acyclic graph} (CPDAG) which includes both directed and undirected edges.% (i.e. structures in the form: $A\rightarrow B \leftarrow C$).

\

\noindent \textbf{Definition 1 (d-separation) \cite{ramsey2012adjacency}}. \textit{In a DAG, a path $p$ between vertices $A$ and $B$ is active (d-connecting, where ``d" stands for dependence) relative to a set of vertices $\mathbf{C}$ $(A,B \notin \mathbf{C}$) if (i) every non-collider on $p$ is not a member of $\mathbf{C}$; (ii) every collider on $p$ is an ancestor of some member of $\mathbf{C}$. Two sets of variables $\mathbf{A}$ and $\mathbf{B}$ are said to be d-separated by $\mathbf{C}$ if there is no active path between any member of $\mathbf{A}$ and any member of $\mathbf{B}$ relative to $\mathbf{C}$.}

\

%An equivalence class of DAGs can be uniquely described by a \textit{completed partially directed acyclic graph} (CPDAG). A partially directed acyclic graph (PDAG) is a graph where the edges are either directed or undirected and one cannot trace a cycle by following the directions of the directed edges and any directions of the undirected edges. A PDAG is \textit{completed} if (1) every directed edge exists also in every DAG belonging to the equivalence class; (2) for every undirected edge, $X_i-X_k$, there exists a DAG with $X_i \leftarrow X_k$ and a DAG with $X_i \rightarrow X_k$ in the equivalence class.
 Let $P$ be the joint probability distribution of $\mathbf{V}$. The following assumptions are set when applying the PC algorithm.

\

\noindent \textbf{Assumption 1.} \textit{(Causal Markov Condition \cite{Spirtes2000}) $G$ and $P$ satisfy the Causal Markov Condition if and only if given the set of all its parents, a node of $G$ is probabilistically independent of all its non-descendents in $G$.}
\\[0.1in]
\noindent
\textbf{Assumption 2.}\textit{ (Faithfulness Condition \cite{Spirtes2000}) $G$ and $P$ satisfy the Faithfulness Condition if and only if  no conditional independence holds unless entailed by the Causal Markov Condition.}
\\[0.1in]
\noindent
\textbf{Assumption 3.} \textit{(Causal sufficiency \cite{Spirtes2000}) For every pair of variables which have their observed values in a given dataset, all their common causes also have observations in the dataset.}

\subsection{The original-PC algorithm}
%The PC algorithm \cite{Spirtes2000},  named after its inventors Peter Spirtes and Clark Glymour,  presents a detailed  implementation of the IC algorithm. The PC algorithm starts from a complete, undirected graph and deletes edges recursively  based on conditional independence decisions. This yields an undirected graph which can then be partially directed and further extended to a \textit{completed partially directed acyclic graphs (CPDAG)}. We present the PC algorithm and its variants in the rest of this section.
The PC algorithm \cite{Spirtes2000} (original-PC algorithm henceforward) has two main steps. In the first step, it learns from data a skeleton  graph, which contains only  undirected edges. In the second step, it orients the undirected edges to form an equivalence class of DAGs.  As the first step of the PC algorithm contributes  to most of the computational costs, we only focus on the modification of this skeleton learning step in this paper, and information about the edge orientation step can be found in \cite{Spirtes2000}.

\

The theoretical foundation of the original-PC algorithm \cite{Spirtes2000} is that if there is no link (edge) between nodes $X$ and $Y$, then there is a set of vertices $\mathbf{Z}$ that either are neighbours of $X$ or  $Y$ such that $X$ and $Y$ are independent conditioning on $\mathbf{Z}$. In other words, $\mathbf{Z}$ disconnects $X$ and $Y$. This foundation is presented as Theorem 1 below.

\

\noindent \textbf{Theorem 1 } \cite{Spirtes2000}. \textit{ If vertices $X$ and $Y$ are not adjacent in a DAG $G$, then there is a set of vertices $\mathbf{Z}$ which is either a subset of $adj(X,G) \backslash {Y}$ or a subset of $adj(Y,G) \backslash {X}$ such that $\mathbf{Z}$ d-separates $X$ and $Y$ in $G$.}

\

Based on Theorem 1, a naive approach for learning the skeleton is that we can search exhaustively for all possible neighbours of $X$ and $Y$ and perform conditional independence tests to justify if there is a set $\mathbf{Z}$ disconnecting $X$ and $Y$. Obviously, this is an inefficient approach, as the neighbours of $X$ and $Y$ are unknown, and thus the search space is all variables other than $X$ and $Y$. In the following we explain how the original-PC algorithm works and list the details of  the skeleton learning step of the original-PC algorithm in Algorithm 1.

\begin{algorithm}[h]
%\SetAlgoNoLine
\KwIn{Dataset $D$ with a set of variables $\mathbf{V}$, and significant level $\alpha$}
\KwOut{The undirected graph $G$ with a set of edges $\mathbf{E}$}
Assume all nodes are connected innitially\\
Let depth $d=0$\\
\Repeat{$|adj(X,G)\backslash \{Y\}|<d$ for every pair of adjacent vertices in $G$}
{
        \For{each ordered pair of adjacent vertices $X$ and $Y$ in $G$}
        {
   	   \If{($|adj(X, G)\backslash \{Y\}| >= d$) }
	   {
        		\For{each subset $Z \subseteq adj(X, G)\backslash \{Y\}$ and $|Z|=d$}
        		{
        			Test $I(X, Y|Z)$\\
			\If {$I(X, Y|Z)$}
			{
				Remove edge between $X$ and $Y$\\
                Save $Z$ as the separating set of $(X, Y)$\\
				Update $G$ and $\mathbf{E}$\\
				break						
			}
		}
      	  }
       }
       Let $d=d+1$
}
\caption{Step 1 of the original-PC algorithm: learning the skeleton}
\label{alg:one}
\end{algorithm}

In the skeleton learning step (Algorithm 1), the orginal-PC algorithm starts with the fully connected network and uses the conditional independence tests to decide if an edge is removed or retained.  For each edge, the PC algorithm tests if the pair of variables connected by the edge,  $X$ and $Y$, are independent conditioning on a subset $\mathbf{Z}$ of all neighbours of $X$ and $Y$. The CI tests are organised by levels (based on the size of the conditioning sets, e.g. the depth $d$). At the first level ($d=0$), all pairs of vertices are tested conditioning on the empty set. Some of the edges would be deleted and the algorithm only tests the remaining edges in the next level ($d=1$). The size of the conditioning set, $d$, is progressively increased (by one) at each new level until $d$ is greater than the size of the adjacent sets of the testing vertices.  The key feature that makes the PC algorithm efficient in sparse true underlying graphs is that the neighbours of a particular node  are dynamically updated when an edge is removed. Therefore, the number of conditional independence tests is small when the true graph is sparse \cite{Spirtes2000}.  If the conditional independence tests are truly correct, then the PC algorithm will return the true graph as stated in Theorem 2 \cite{Spirtes2000}.%Figure 3.4 shows an example of the PC algorithm for 4 nodes $A,B,C,D$. More details of the PC algorithm can be found in \cite{Spirtes2000, Kalisch2007}.
\\[0.1in]
\noindent \textbf{Theorem 2} \cite{Spirtes2000}.\textit{  Let the distribution $P$ be faithful to a DAG $G = (\mathbf{V};\mathbf{E})$, and assume that we are given perfect conditional independence information about all pairs of variables $(X,Y)$ in $\mathbf{V}$ given subsets $S$. Then the output of the PC-algorithm is the CPDAG that represents $G$.}
\\[0.1in]
\indent
Figure 1 (top) shows an example of original-PC algorithm being applied to a dataset with four nodes, $A, B, C,$ and $D$. The original-PC algorithm starts with the fully connected graph. At the first level, all edges are tested conditioning on the empty set. In the example, after level 1 tests, there are three edges left. At the next level, each remaining edge will be tested conditioning on each neighbour (size 1) of the testing variables. For example, with the edge between $A$ and $B$ we have at most two tests which are conditioning on $C$ and conditioning on $D$. If the test returns independence (e.g. $I(A,B|C)$), we remove the edge from the graph and move to testing the other edge. The procedure will stop when there is no test to perform.
\begin{figure}[h]
\begin{minipage}[htbp]{0.5\textwidth}
\includegraphics[width= 3.5in, height=3in]{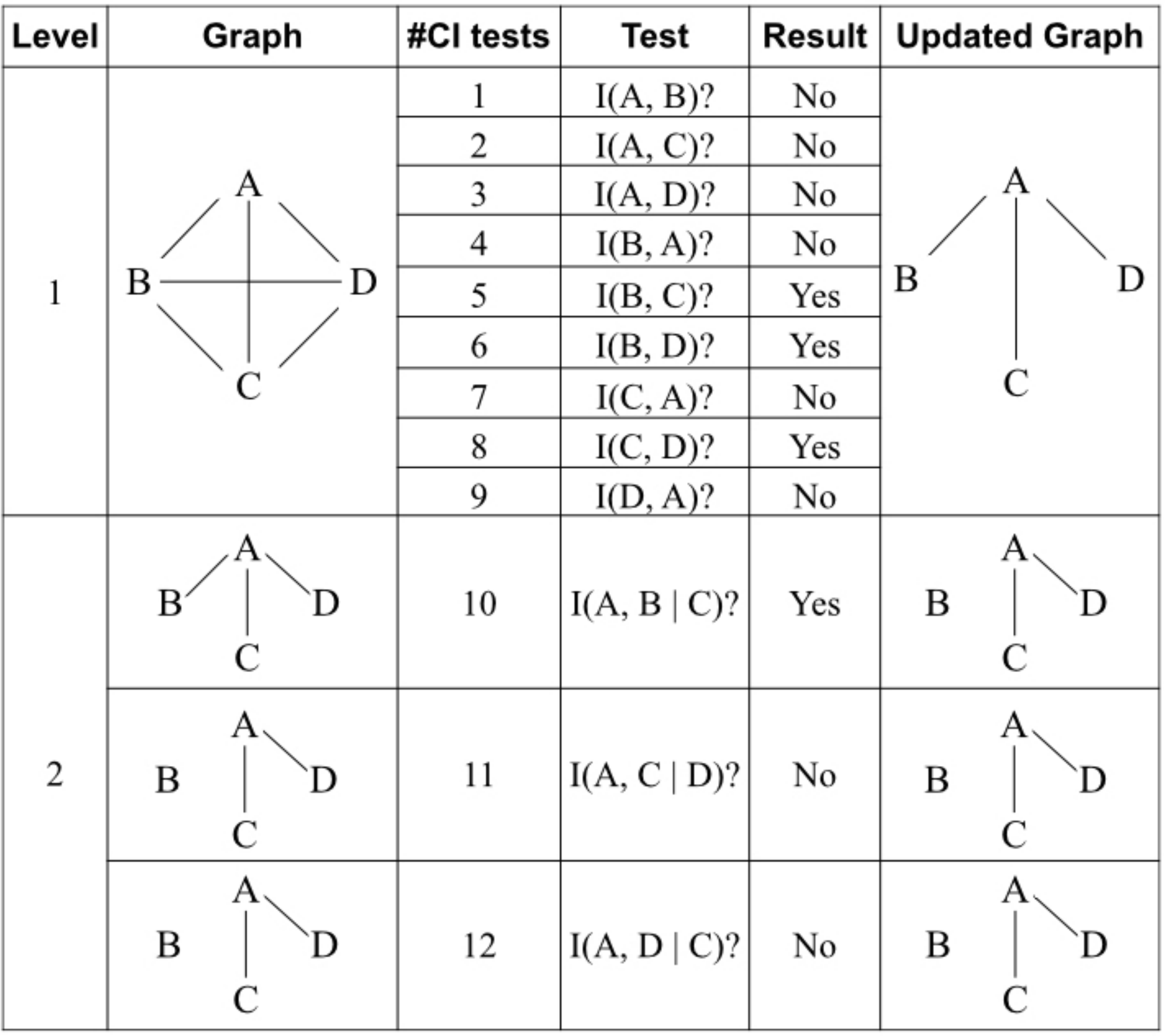}
% where an .eps filename suffix will be assumed under latex,
% and a .pdf suffix will be assumed for pdflatex; or what has been declared
% via \DeclareGraphicsExtensions.
\end{minipage}
\begin{minipage}[htbp]{0.5\textwidth}
\includegraphics[width= 3.5in, height=3.5in]{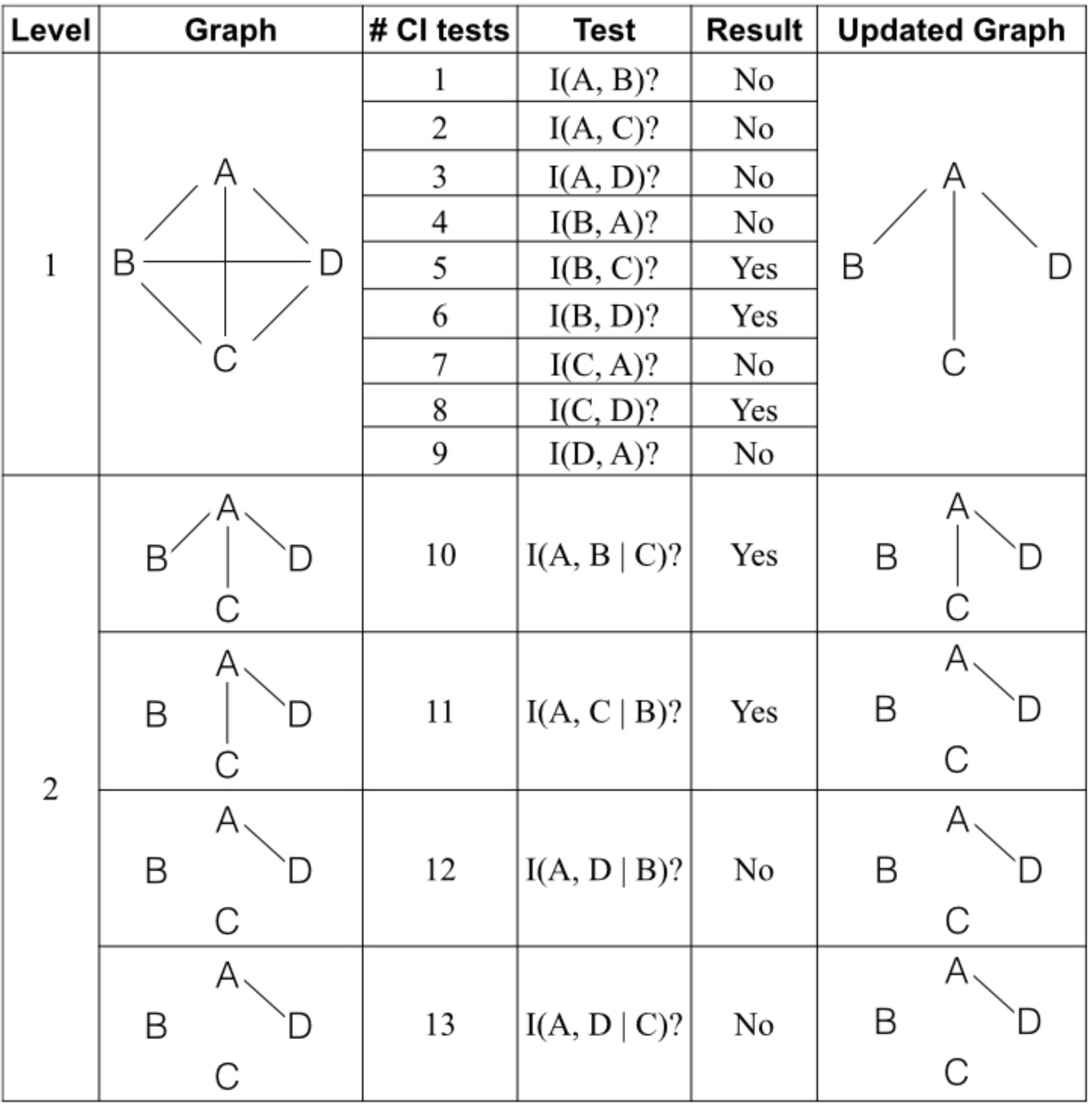}

\end{minipage}
\caption{Examples of applying  PC  (top) and  stable-PC  (bottom).}
\label{PC}
\end{figure}

\subsection{The stable-PC algorithm}
The CI tests in the original-PC algorithm are prone to mistakes as the number of samples is limited. Moreover, incorrectly removing or retaining an edge would result in the changes in the neighbour sets of other nodes, as the graph is updated dynamically. Therefore, the output graph is dependent on the order in which we perform the conditional independence tests. In other words, given a lexical order of performing the tests, the order of the variables in the input data will affect the output graph. In Figure 1 (top), at the beginning of level 2, after we perform test $\#10$, the edge between $A$ and $B$ is removed and the graph is updated. This also means that the adjacent set of $A$ is also updated, i.e. $adj(A, G)=\{C, D\}$. Therefore, when we test the edge between $A$ and $C$, we do not need to condition on $B$. The updates make the original-PC algorithm efficient, but they also cause problems when the CI tests are making mistakes. If test $\#10$ wrongly removes the edge between $A$ and $B$, then we miss the test $I(A,C|B)$ which may remove the edge between $A$ and $C$.  Moreover, if we test $I(A,C|B)$ first and remove the edge between $A$ and $C$, we will end up with a different graph. In other words, when we change the order of the variables in the dataset, e.g. $A, C, B,$ and $D$, the result of the original-PC algorithm may change.

\

Colombo et al. \cite{colombo2012} proposed a modification to the original-PC algorithm to obtain a stable output skeleton  which does not depend on how variables are ordered in the input dataset. In this method (called stable-PC algorithm), the neighbour (adjacent) sets of all nodes are queried and kept unchanged at each particular level. As a result, an edge deletion at one level does not affect the conditioning sets of  the other nodes, and thus the output is independent with the variable ordering. For instance, at the beginning of level 2 in the above example, the adjacent set of node $A$ is $adj(A, G)=\{B, C, D\}$. This adjacent set is kept unchanged at this level. After performing test $\#10$, we remove the edge between $A$ and $B$ in the graph, but the $adj(A, G)$ is not updated. Therefore, we still consider $B$ as an adjacent node of $A$, and  tests $\#11$ and $\#12,$ still need to be performed as shown in Figure 1 (bottom). Test $\#11$ removes the edge between $A$ and $C$, and therefore generates different output graph compared to the original-PC algorithm.

\

However, this modification requires performing more conditional independence tests in each level of the algorithm, and thus it increases further the runtime of the algorithm. In the following section, we adopt this modification to design a modified PC algorithm which is both efficient and order-independent.
\section{Parallel-PC algorithm}

\subsection{Parallel Computing: Potentials and Challenges}

 Computer hardware technology has been advanced significantly in the last decade. Nowadays, it is common to have personal desktop computers and even laptops have 4 cores in their CPUs.  Likewise, the number of cores in today's servers has jumped to 8, 16, 32 or more. These resources provide a great potential of utilising parallel computing for important algorithms, including the PC algorithm, for personal computers. In fact, the majority of end users are using their personal computers for doing research. Therefore, parallel algorithms would have significant impact if they work well for personal computers and do not require users' technical knowledge about parallelism and high-performance clusters.

\

 The idea of parallelism is that we break down a big task into several different smaller subtasks and distribute them over different cores of the computer's CPU to run in parallel. The results from all subtasks will then be integrated to form the result of the original task. Theoretically, this divide-and-conquer approach can speed up an algorithm by the maximum  number-of-core times. This approach, however, requires the subtasks to be unrelated. In other words, the result from each subtask must be independent of one another and no communication is allowed between them.

\

The  independence requirement has been a great barrier for many sequential algorithms, including the stable-PC algorithm, to be parallelised. In the stable-PC algorithm, the performance bottleneck is the  huge number of CI tests, which can reach hundreds of millions in real-world datasets. The CI tests, however, are order-dependent across different levels. Since the stable-PC algorithm updates the adjacent sets of all nodes at each level after all the CI tests of the level are completed, the test results of a particular level will influence the results of the next level. Therefore, it seems infeasible to parallelise the stable-PC algorithm as a whole.

\subsection{The parallel-PC Algorithm}

Our proposed strategy is to parallelise the CI tests, not across different levels, but inside each level of the stable PC-algorithm. This approach is feasible because the CI tests at a particular level are independent of each other. Since the graph is only updated at the end of each level, the result of one CI test does not affect the others. Therefore, the CI tests at a level can be executed in parallel without changing the final result. Furthermore, this approach also enjoys the advantage of knowing the number of CI tests of each level in advance. This allows the CI tests to be evenly distributed over different cores, so that the parallelised algorithm can achieve maximum possible speedup.

\begin{algorithm}[!h]
%\SetAlgoNoLine
\KwIn{Dataset $D$, significant level $\alpha$, $P$ cores, memory-efficient indicator $s$, number of edges per batch $t_b$} %($= f_m / 2$ by default, where $f_m$ is the free memory in MBs)}
\KwOut{The undirected graph $G$ with a set of edges $E$}
 Assume all nodes are connected in graph $G$ \\
Let depth $d = 0$\\
\Repeat{$|adj(X, G) \backslash \{Y\}| < d$ for every pair of adjacent vertices in $G$}
{
	Query and fix the adjacent set $adj(X, G)$ of each node $X$ in $G$\\
	Compute the set $J$ of unordered pairs of adjacent vertices $(X, Y)$ in $G$\\
	%Split $J$ into $P$ evenly sized subsets: $J = \{J_i\}_{i=1}^P$\\
	\tcp{Parallelisation Step}
	\For{each batch of $t_b$ edges ($t_b = |J|$ if $s = FALSE$)}
	{
	Distribute the edges in the batch evenly into $P$ cores, each with $J_p$ edges\\
	\For{each core $p = 1 \dots P$ in parallel}
	{
		\For{each pair $(X, Y) \in J_p$}
		{
			Let $k_{X, Y}^p$ indicate if $(X, Y)$ is adjacent, initialize $k_{X, Y}^p = \text{TRUE}$\\
			\tcp{On $X$'s neighbours}
			\If{$|adj(X, G) \backslash \{Y\}| \ge d$}
			{
				\For{each subset $Z_X \subseteq adj(X, G) \backslash \{Y\}$ and $|Z_X| = d$}
				{
					%Test $I(X, Y | Z_X)$\\
					\If{$I(X, Y | Z_X)$}
					{
						$k_{X, Y}^p = \text{FALSE}$\\
						\textbf{break}
					}
				}
			}
			\tcp{On $Y$'s neighbours}
			\If{$|adj(Y, G) \backslash \{X\}| \ge d$}
			{
				\For{each subset $Z_Y \subseteq adj(Y, G) \backslash \{X\}$ and $|Z_Y| = d$}
				{
					%Test $I(X, Y | Z_Y)$\\
					\If{$I(X, Y | Z_Y)$}
					{
						$k_{X, Y}^p = \text{FALSE}$\\
						\textbf{break}
					}
				}
			}
		}
	}
	\tcp{Synchronisation Step}
	\For{each core $p = 1 \dots P$}
	{
		\For{each pair $(X, Y) \in J_p$}
		{
			\If{$k_{X, Y}^p = \text{FALSE}$}
			{
				Remove the edge between $X$ and $Y$ and update $G$ and $E$\\
			}
		}
	}
	}
	Let $d = d + 1$\\
}
\caption{The parallel-PC algorithm}
\label{alg:one}
\end{algorithm}

\

Within a level (depth) of Algorithm 2, we employ the parallel computing paradigm  to parallelise their executions (the parallelisation step). At each level with the conditioning sets of size $d$, the sequential process in the stable-PC algorithm is replaced by a three-staged process: (1) the CI tests are distributed evenly among the cores, (2) each core performs its own sets of CI tests in parallel with the others, and (3) the test results in all cores are integrated into the global graph. This three-staged process is applied at all the levels (the depth $d$) of the algorithm.

\

\indent We observed that it is not efficient to distribute the CI tests of the same edge to different cores. Given an edge between $X$ and $Y$, we need to perform the CI tests conditioning on the variables in  $adj(X, G)$ and in $adj(Y, G)$. However, if we firstly perform the CI tests conditioning on variables in $adj(X, G)$ and the edge between $X$ and $Y$ is removed, then the CI tests conditioning on variables in $adj(Y, G)$ are unnecessary. We resolve this dependency by grouping the CI tests of the same edge together, instead of separating them as in the stable-PC algorithm. In Algorithm 2, if the CI tests between $X$ and $Y$  return independence when conditioning on $\mathbf{Z}\subset adj(X, G)\backslash \{Y\}$, the algorithm will not perform the CI tests conditioning on variables in $adj(Y, G)$. Grouping those tests together, the algorithm ensures the CI tests of the same edge will not be distributed to different cores of the CPU, and thus reduce the number of unnecessary CI tests that need to be performed.

\

In Algorithm 2, we provide an optional input parameter, memory-efficient indicator $s$. If $s$ is set to TRUE, the algorithm will detect the free memory of the running computer to estimate the number of edges in $G$ that will be distributed evenly to the cores. This step is to ensure that each core of the computer will not hold off a big amount of memory while waiting for the synchronisation step. The memory-efficient procedure may consume a little bit more time. However, this option is recommended for computers with limited memory resources or for big datasets.

\subsection{Implementation}

We implement our parallel-PC algorithm using $R$ \cite{R}. Our implementation is based on the stable-PC algorithm in the R package \textit{pcalg}  \cite{pcalg} and the native R library \textit{parallel}. The implementation has been tailored to work in Linux,  MacOS, and Windows operating systems. The codes of parallel-PC with and without the memory efficient option together with the instructions of running the algorithm are available at: http://nugget.unisa.edu.au/ParallelPC

\section{Experimental evaluation and applications}
\subsection{Datasets}

\begin{table}[!h]
%% increase table row spacing, adjust to taste
\renewcommand{\arraystretch}{2.0}
% if using array.sty, it might be a good idea to tweak the value of
% \extrarowheight as needed to properly center the text within the cells
\caption{Real world gene expression datasets}
\label{datasets}
\centering
%% Some packages, such as MDW tools, offer better commands for making tables
%% than the plain LaTeX2e tabular which is used here.
\begin{tabular}{|c|c|c|c|c|}
\hline
Dataset & \#samples & \#variables& stable-PC runtime\\
\hline
NCI-60  &47 & 1190 & 13.56 mins\\
\hline
MCC & 88 & 1380 & $~$ 59.87 mins \\
\hline
BR51 & 50&1592& 68.37 mins\\
\hline
\textit{S. cerevisiae}&63&5361& 207.41 mins\\
\hline
\textit{S. aureus} &160 &2810& 477.20 mins\\
\hline
DREAM5-Insilico&805&1643&$>$ 24 hours \\
\hline
\end{tabular}
\end{table}

%& \cite{le2013,Le2014,le2013b}
%&  \cite{Le2014}
%& \cite{Le2014}
%& \cite{Maathuis2010}
%& \cite{marbach2012}
%&\cite{marbach2012}
In this section, we evaluate the efficiency of the proposed algorithm by applying it to five real world datasets and a synthetic dataset (see Table 1 for a summary of the datasets). The first three datasets are commonly used in research of inferring the regulatory relationships between microRNAs (a type of gene regulators) and genes. These datasets were downloaded from \cite{Le2014}. Meanwhile, the last three datasets are used for discovering the regulatory relationships between transcription factors (another type of gene regulators) and the target genes regulated by them. The  \textit{S. cerevisiae} dataset was provided by the authors of \cite{Maathuis2010}, and the last two datasets were downloaded from \cite{marbach2012}. The number of samples and variables in each dataset together with the runtime of the stable-PC algorithm are summarised in Table 1. We choose these datasets for the experiments, as they require a wide range of runtime when applying the stable-PC algorithm. All of the datasets are available at: nugget.unisa.edu.au/ParallelPC

\subsection{Efficiency evaluations}
\begin{figure*}[t]
\centering
\includegraphics[width=6in, height=4in]{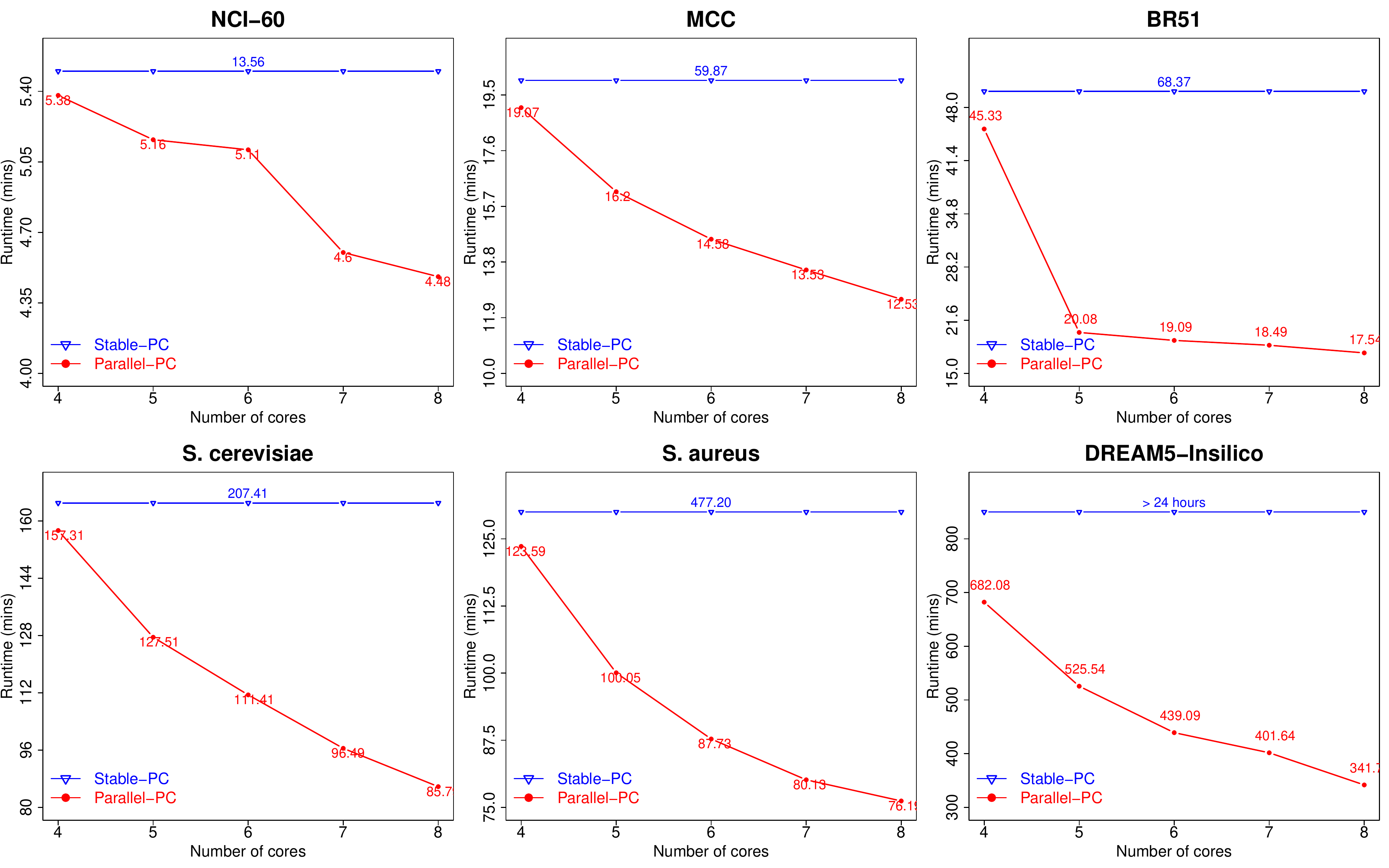}
% where an .eps filename suffix will be assumed under latex,
% and a .pdf suffix will be assumed for pdflatex; or what has been declared
% via \DeclareGraphicsExtensions.
\caption{Runtime of the algorithms on the six datasets. We run each experiment three times, and each value in the figure is the average runtime of the three runs. The average variation coefficient (standard deviation / mean) of the PC-parallel algorithm in the three runs is 0.039 and the maximum variation coefficient is 0.055.}
\label{EMT}
\end{figure*}
 We apply the stable-PC algorithm, and the parallel-PC algorithm to all of the datasets and compare their performance in terms of the runtime.  For each dataset, we run five versions of the parallel-PC algorithm using 4, 5, 6, 7, and 8  cores CPU respectively. Meanwhile, for the stable-PC algorithm, we used the implementation in the \textit{pcalg}\cite{pcalg} $R$ package. We choose the range from four to eight cores for the experiments, as a modern personal computer normally has four cores and a small server usually has eight cores. We run all the experiments on a Linux server with 2.7 GB memory and 2.6 Ghz per core CPU. All experiments were run three times and the average value of the three runs was reported.

\

 As shown in Figure 2, the runtime of the parallel-PC algorithm is much shorter compared to the stable-PC algorithm. Please note that the parallel-PC algorithm generates the results that are independent of the order of the variables in the datasets, and the output results are consistent with the stable-PC algorithm. %Although the resulting graphs from the parallel-PC and the PC-stable are not the same as the skeleton generated by the original PC algorithm which is order-dependent on the variables in the dataset, we add the runtime of the original PC algorithm in the figures for references.

\

 In the relatively small datasets, NCI-60,  stable-PC took around 14 minutes to complete, but parallel-PC  with 4-core CPU only spent around 5 minutes. For this small dataset, increasing the number of cores does not reduce the runtime significantly due to the cost of synchronisation between the cores.

\

 For the BR51 and MCC datasets, it took more than one hour each for running the stable-PC algorithm. The 6-core parallel-PC algorithm only used less than one third (less than 20 minutes) of its counterpart's runtime. Meanwhile, there is a clear trend in the \textit{S. cerevisiae} dataset that the more powerful CPU (up to 15 cores) we use,  the shorter runtime the parallel-PC algorithm has.

\

 The efficiency of the parallel-PC algorithm is even higher with big datasets. In the S.aureus dataset, the runtime of the stable-PC algorithm was around 8 hours. Meanwhile, using a 4-core version of the parallel-PC algorithm, it took only around two hours. This is a significant result for end users who only use personal computers for doing their research experiments. The runtime is further reduced when using computer CPU with more cores. For instance, it took just more than one hour (76.1 minutes) for running this dataset with an 8-core CPU computer (server). With the DREAM5-Insilico dataset, after more than 24 hours the stable-PC still could not produce any results, and we had to abort the job. This is a typical situation when applying the stable-PC algorithm to big datasets, as we may not be able to predict its running time on a real world dataset. However, our 4-core parallel-PC algorithm managed to finish within around 12 hours, and the 8-core version took less than 6 hours (341.2 minutes).

\

 The experiment results suggest that our proposed algorithm has significantly shorter runtime than the stable-PC algorithm. Given that our method generates the same output and requires no extra effort from users compared to that of the stable-PC algorithm, the parallel-PC is more useful in exploring causal relations in high dimensional data, especially gene expression data. Note that the provided software tool is not restricted to only personal PCs, and it is ready for clusters or super computers with CPUs of more cores. To provide more insight into the experiments in Figure 2, we run the experiments on a super computer (with a 48-core CPU) using up to 16 cores of the CPU and the results are in Figure 3. For most of the datasets, the runtime was not improved further when we used more than 14 cores of the CPU.

\begin{figure*}[!t]
\centering
\includegraphics[width=6in, height=4in]{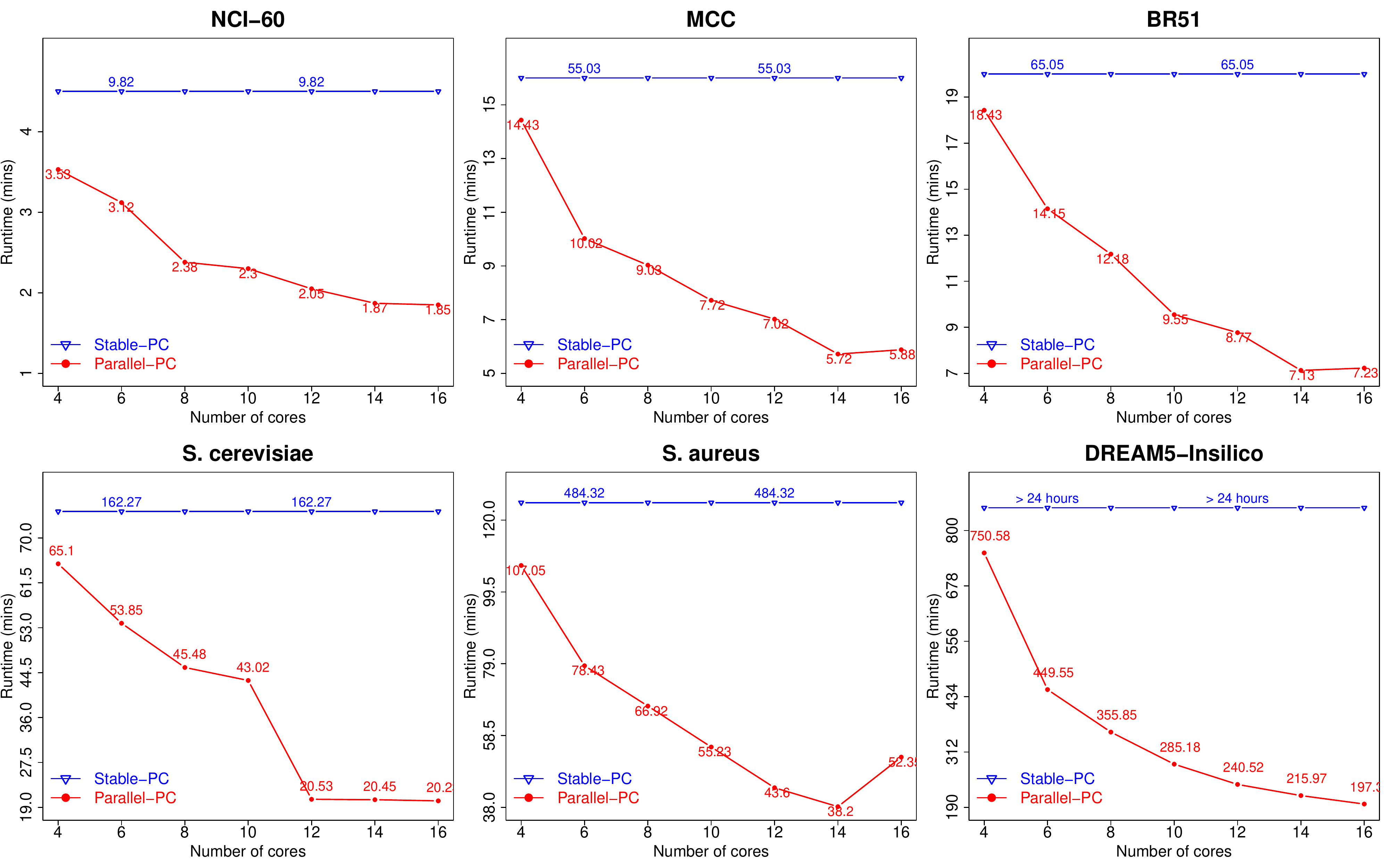}
% where an .eps filename suffix will be assumed under latex,
% and a .pdf suffix will be assumed for pdflatex; or what has been declared
% via \DeclareGraphicsExtensions.
\caption{Runtime of the algorithms using up to 16 cores with the six datasets.}
\label{morecores}
\end{figure*}

\subsection{Scalability}

\begin{figure}[!t]
\centering
\begin{minipage}[htbd]{0.5\textwidth}
\includegraphics[width=3.0in, height=2.5in]{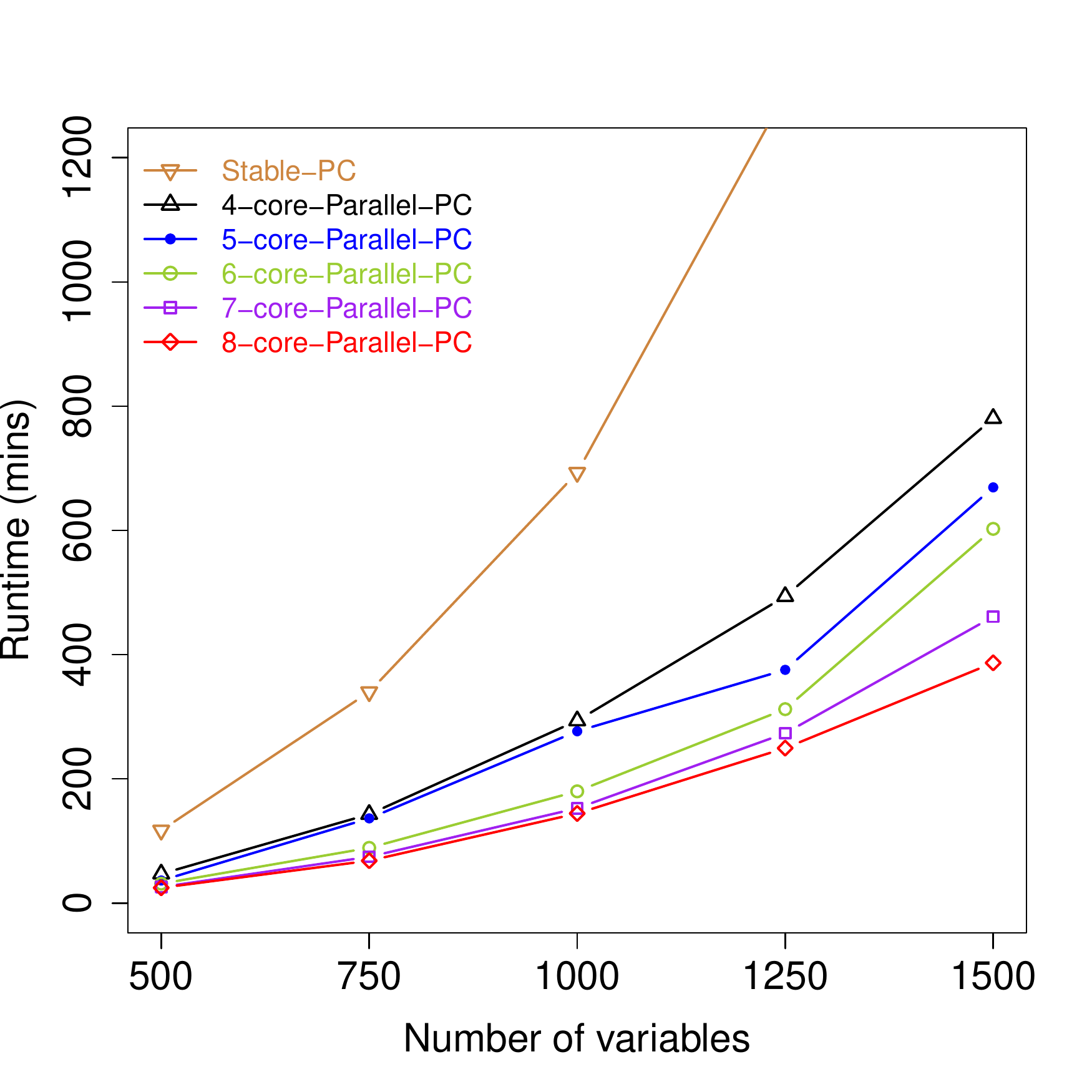}
\end{minipage}
\begin{minipage}[htbd]{0.5\textwidth}
\includegraphics[width=3in, height=2.5in]{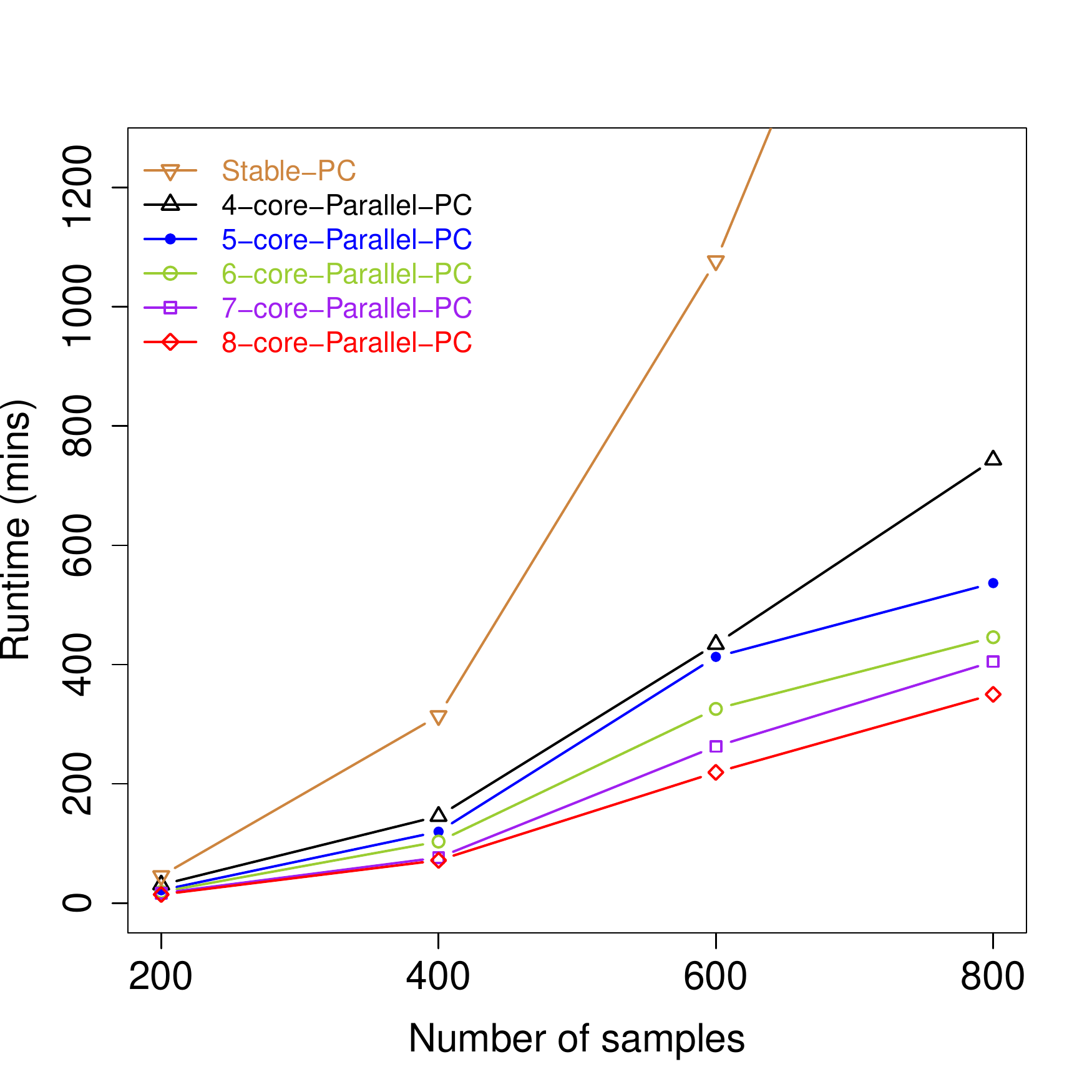}
\end{minipage}
\caption{Scalibility of the parallel-PC algorithm with number of attributes (top) and with data size (bottom).}
\label{fig_sim}
\end{figure}
To compare the scalability of the parallel-PC algorithm and the stable-PC algorithm, we used the DREAM5-insilico dataset as a base and randomly sample from it four datasets with different numbers of samples (200, 400, 600, and 805) and five datasets with different numbers of variables (500, 750, 1000, 1250, and 1500).

\

Figure 4 (top) shows that the runtime of the stable-PC algorithm increases significantly when  the number of variables is increased in a dataset. The reason is that the number of CI tests is increased when the number of variables is increased. The parallel-PC algorithm suffers from the same problem, but the increasing trend of the runtime is less significant compared to that of the stable-PC algorithm.

\

Similarly, the runtime of both PC and parallel-PC goes up when we increase the number of samples in the dataset as shown in Figure 4 (bottom). However, the reason for runtime increase is not because of the increased number of CI tests, but due to the increase in the time taken to perform a test. When we increase the number of samples in the input dataset, the time for performing one CI test is increased, and thus resulting in the large increase in total runtime given that the number of CI tests is huge.

\subsection{Impact of the memory-efficient option}
As mentioned in Section 4, the memory-efficient version of the algorithm uses less amount of memory  at any time point while running the algorithm. We observed that peak memory usage in the algorithm with memory-efficient option ($s$=TRUE) is about half of that of the algorithm without using the option across the six datasets in Table 1. However, there is a trade-off between memory efficiency and time efficiency as show in Table 2. The experiments were conducted on a MacBook Pro with 8GB RAM, 2.6 GHz. Running \textit{S. cervisiae} with $s=FALSE$ failed to finish due to memory exhaustion.

\begin{table}[!h]
\renewcommand{\arraystretch}{2.0}
\caption{Impact of the memory-efficient option}
\label{datasets}
\centering
\begin{tabular}{|c|c|c|c|c|}
\hline
\multirow{2}{*}{Dataset}& \multicolumn{2}{c|}{Time (mins)}& \multicolumn{2}{c|}{Memory (MBs)}\\
%\hline
 & $s=F$ &$s=T$ &$s=F$ &$s=T$\\
 \hline
NCI-60 & 2.2& 2.9&919&421.5\\
MCC & 9.4 &11.6&1323&533.5\\
BR51&11.9&15.3&1593&679.4\\
Insilico&531.7&557.9&1533&784\\
\textit{S. aureus}&92.1&91.3&4038&1857.1\\
\textit{S. cerevisiae}&NA&91.3&$>$8000&6004\\
\hline
\end{tabular}
\end{table}

\iffalse
\begin{table}[!h]
\renewcommand{\arraystretch}{1.0}
\caption{Impact of the memory-efficient option}
\label{datasets}
\centering
\begin{tabular}{|c|c|c|c|c|c|c|c|}
\hline
& & NCI-60&MCC&BR51&Insilico&\textit{S. aureus} &\textit{S. cerevisiae}\\
\hline
\multirow{2}{*}{$s=FALSE$} & Time(mins)& 2.2& 9.4& 11.9&531.7&92.1&NA\\
%\hline
&Memory (MBs) & 919 &1323 & 1593 & 1533&4038&$>8000$\\
\hline
\multirow{2}{*}{$s=TRUE$} &  Time(mins)& 2.9 & 11.6 & 15.3&557.9&91.3&91.3\\
%\hline
&Memory (MBs) & 421.5& 533.5&679.4&784 &1857.1&6004.4\\
\hline
\end{tabular}
\end{table}
\fi

\section{Application in inferring miRNA-mRNA regulatory relationships}
\subsection{Computational methods for identifying miRNA targets}
microRNAs (miRNAs) are important gene regulators at post-transcriptional level. They regulate gene expression by promoting messenger RNA (mRNA) degradation and repressing translation \cite{Bartel2004}. miRNAs control a wide range of biological processes and are involved in several types of cancers \cite{Bartel2004, KimVN2006}.  In the last decade, predictions of miRNA functions through identifying miRNA-mRNA regulatory relationships by computational methods have increasingly achieved promising results. Computational approaches are proving to be effective in generating hypotheses to assist with the design of wet-lab experiments for confirming miRNA targets.

\

Several methods have been proposed to identify miRNA-mRNA regulatory relationships. In the first stream, researchers identified miRNA-mRNA interactions using sequence data \cite{enright2004, Lewis2005,Krek2005}. Although these methods can predict  potential miRNA targets, the results may contain a high-rate of  false positives and false negatives  \cite{Rajewsky2006}. In the second stream of research, various computational methods have been devised  to use expression profiles in the study of  miRNA-mRNA regulatory relationships.  The principle of these methods is to investigate if a change in the miRNA expression level would result in a change in the mRNA expression. Some highlights of the methods are  correlation analysis \cite{liuH2010, van2010}, regression models \cite{ lu2011, muniategui2012}, population-based probabilistic learning model \cite{Joung2007}, rule based methods \cite{Tran2008}, Bayesian network learning \cite{Huang2007, liu2009a, le2013b}, causal inference techniques  \cite{le2013, Zhang2014a, Zhang2014b}, and ensemble methods \cite{Le2015}. Please refer to \cite{Le2014} for a survey of different miRNA target prediction methods.

\subsection{Inferring miRNA-mRNA regulatory relationships with efficient-IDA}

In this section, we propose to use parallel-PC to modify the the causal inference method \cite{le2013} for inferring miRNA-mRNA regulatory relationships. The method is based on a causal inference approach called IDA (Intervention effect when the DAG is absent) \cite{Maathuis2009}. IDA firstly learns the causal structure from data using the PC algorithm, it then uses \textit{do-calculus} to estimate the causal effect that a variable has on the other.  The estimated causal effects simulate the effects of  randomised controlled experiments. Although the method is proved to be effective in uncovering miRNA-mRNA regulatory relationships, it has high computational complexity and suffers from the order-dependent problem. In this section, we propose to tackle these problems by modifying the IDA-based method by using parallel-PC  rather than PC as the first step of the IDA algorithm. The modified algorithm is called efficient-IDA and includes the following steps.

\begin{itemize}
\item Step 1: Find miRNAs and mRNAs that are differentially expressed between different conditions, e.g. normal vs cancer. The expression profiles of those miRNAs and mRNAs will be combined into a dataset where rows are samples and columns are variables (miRNAs and mRNAs). In our experiments, we download from \cite{Le2014} three processed cancer datasets with differentially expressed profiles.
\item Step 2: Apply parallel-PC to learn the causal structure from data. parallel-PC will help tackle the efficiency and order-dependency problems of the original PC algorithm.
\item Step 3: Use \textit{do-calculus} to estimate the causal effects that the miRNAs have on the mRNAs. For more information on this step, please refer to \cite{le2013}.
\end{itemize}

 The running time of IDA is quite similar to the PC algorithm, as the time for the inference step (Step 3 as described above) is small. Therefore we do not compare the runtime of efficient-IDA and IDA again and please refer to Figures 2 and 3 for details. In the following sub-sections, we compare the performance of efficient-IDA and IDA in correctly identifying miRNA-mRNA regualtory relationships.

\subsection{efficient-IDA discovers more experimentally miRNA targets than IDA}

We applied efficient-IDA and IDA to NCI-60, MCC, and BR51 and compare their performance on discovering the experimentally confirmed miRNA targets.

\

To validate the target predictions from the two methods, we used the combination of four different experimentally confirmed miRNA target databases as the ground truth.  They are Tarbase v6.0 \cite{vergoulis2012}, miRecords v2013 \cite{xiao2009}, miRWalk v2.0 \cite{dweep2011}, and miRTarBase v4.5 \cite{HsuSD2014}. Tarbase, miRecords, and miRTarBase include verified interactions that are manually curated from the literature, and miRWalk contains both the predicted and the experimentally validated interactions. Respectively for Tarbase,  miRecords, miRWalk, and miRTarBase, we have 20095 interactions with 228 miRNAs, 21590 interactions with 195 miRNAs,  1710 interactions with 226 miRNAs, and 37372 interactions with 576 miRNAs. After removing the duplicates, we have in total 62858 unique interactions to be used in the validation.

\begin{figure}[!t]
\begin{minipage}[htbd]{0.5\textwidth}
\includegraphics[width=3in, height=2in]{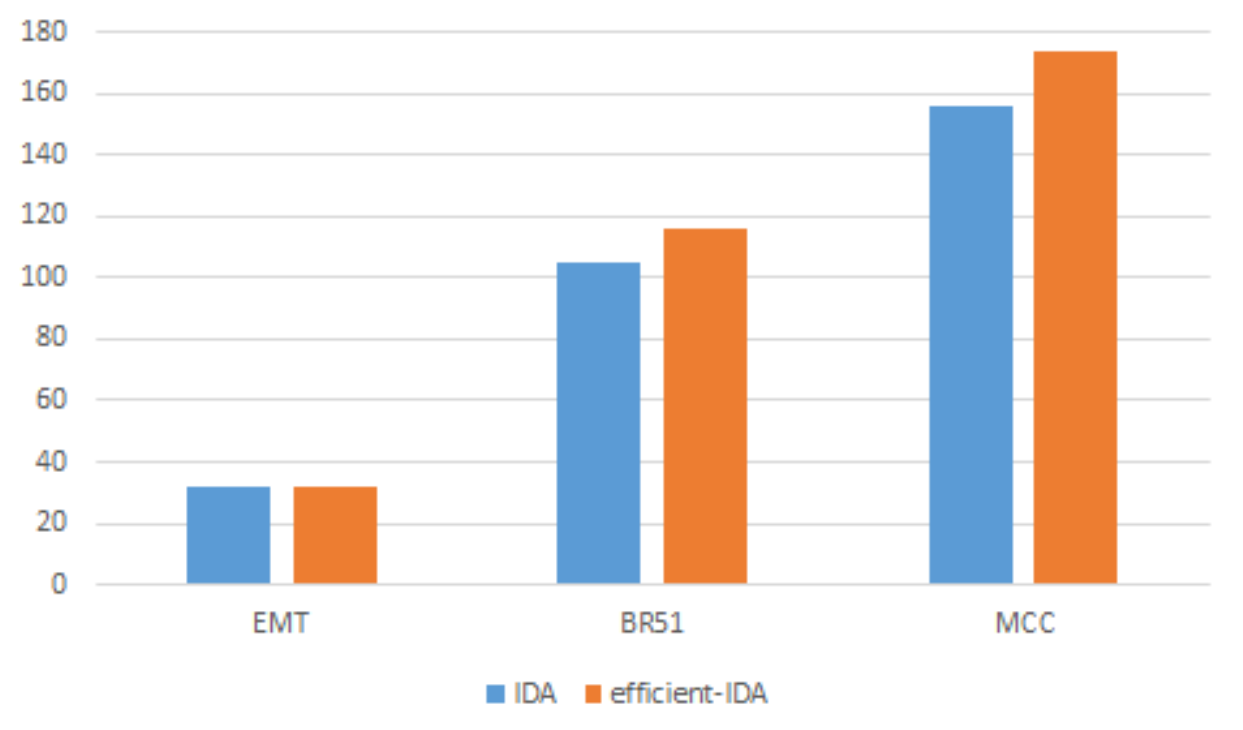}
\end{minipage}
\begin{minipage}[htbd]{0.5\textwidth}
\includegraphics[width=3in, height=2in]{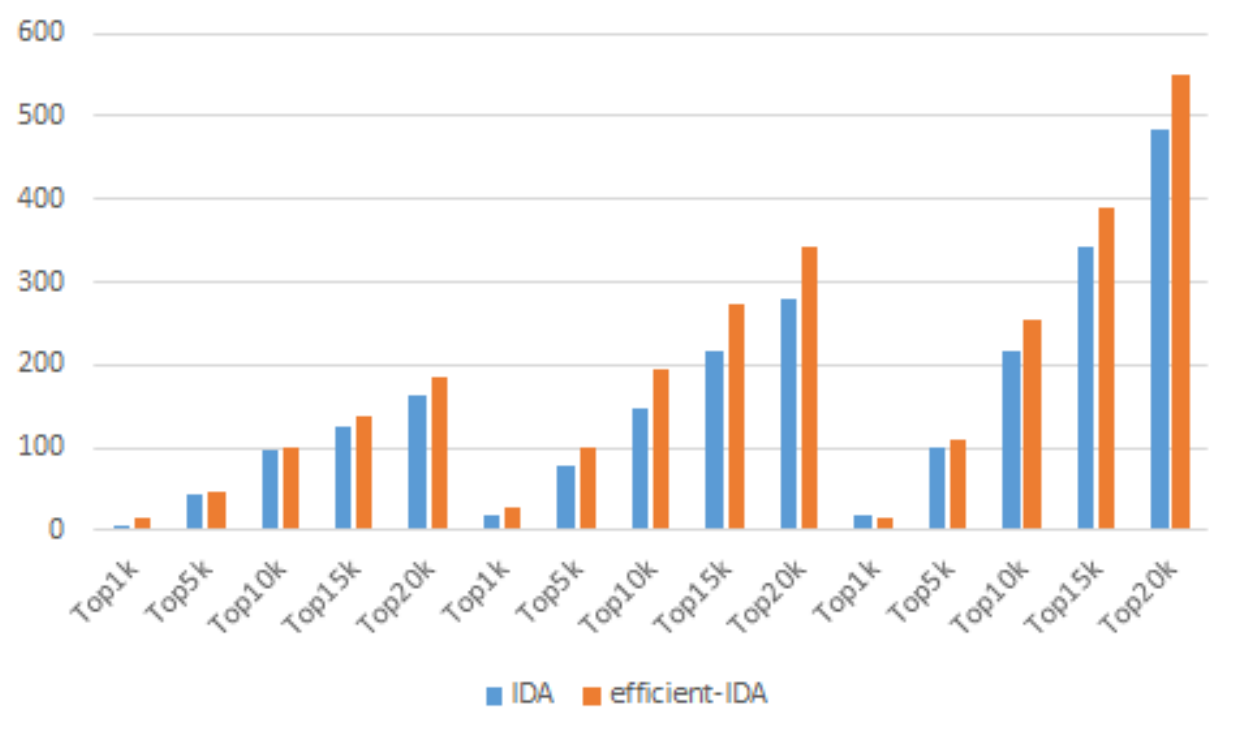}
\end{minipage}
\caption{Number of confirmed miRNA targets identified by IDA and efficient-IDA. The top figure: the top 100 targets of each of the miRNAs in a dataset are extracted for validation. The bottom figure: the top 1k, 5k, 10k, 15k, and 20k interactions in each dataset (from left to right, NCI-60 (EMT), BR51, and MCC) are extracted respectively for validation. }
\label{fig_sim}
\end{figure}

%\begin{figure*}[!t]
%\centering
%\includegraphics[scale=0.5]{top100compare.png}
%% where an .eps filename suffix will be assumed under latex,
%% and a .pdf suffix will be assumed for pdflatex; or what has been declared
%% via \DeclareGraphicsExtensions.
%\caption{Number of confirmed miRNA targets of IDA and efficient-IDA.}
%\label{top100}
%\end{figure*}

\

We used two different ways to extract prediction results for the comparison. Firstly, we extracted the top 100 predicted targets of each of the miRNAs in a dataset and validate them against the experimentally confirmed miRNA targets. Secondly, we extracted the top 1000, 2000, 5000, and 10000 predicted miRNA-mRNA interactions of each method for validation.

\

Figures 5  shows the comparison results of the number of confirmed miRNA targets of efficient-IDA and IDA using the two validation methods respectively. We can see from the figure that efficient-IDA discovers more confirmed miRNA targets than IDA in both of the cases in all datasets. The experiment results suggest that efficient-IDA is not just faster in runtime, but the order-independent property of efficient-IDA helps improve the accuracy of the causal inference method (IDA).

%\subsection{efficient-IDA and IDA discover different set of miRNA targets}

%Venn Diagram
\section{Conclusions and future work}
It is very important to explore causal relationships in real world high dimensional datasets. For instance, the discovered causal relationships can help elucidate the causes of fatal diseases from gene expression data. However, existing causal discovery method do not provide practical solutions to the problem. In this paper, we have developed an efficient algorithm for exploring causal relationships in high dimensional datasets based on the stable-PC algorithm and parallel computing framework.  We modify the stable-PC algorithm and group the CI tests in the algorithm to enable the tests to be performed in parallel. Our method produces the causal relationships that are consistent with the stable-PC algorithm, but much more efficiently regarding the runtime. The experiment results from a wide range of commonly used real world datasets suggest the efficiency of the proposed method. parallel-PC also shows its usefulness by improving the efficiency and accuracy of IDA in inferring miRNA-mRNA regulatory relationships.

% Theoretically, the proposed method can be used in any kind of datasets other than high dimensional data. However, we observed that for datasets with a very large number of samples (tens of thousands), the parallel-PC algorithm is not very efficient. The reason is that it requires a large amount of memory and runtime due to the communication between the cores. However, real world high dimensional datasets, e.g. gene expression datasets, normally have small number of samples (less than thousands of samples), and thus the proposed algorithm is very efficient. In the future, we will tackle this issue, so that the application of the method can be extended to a wider range of datasets.
\

The effectiveness of the parallel-PC algorithm could be further improved with a job weighting and scheduling scheme. In the current version of parallel-PC, we divide the number of CI tests equally and distribute to different cores of the CPU. However, there is no guarantee that the execution time of different cores is the same. A smart job weighting and scheduling scheme would help speed up the algorithm.

\

Finally, a straightforward but useful future work is to apply the parallel-PC algorithm to other PC related causal discovery and causal inference methods. Most of the  constraint based methods are based on the idea of the PC algorithm directly or indirectly. The FCI \cite{Spirtes2000} algorithm and RFCI \cite{colombo2012learning} are based on the PC algorithm to learn causal graphs which allows latent variables. The CCD algorithm \cite{richardson1996discovery} for learning Markov equivalent class  utilises the PC algorithm as the first step. The parallelised versions of these algorithms would help bridging the gap between computer science theory and their applications in practice.

\ifCLASSOPTIONcompsoc
  \section*{Acknowledgments}
\else
  \section*{Acknowledgment}
\fi

This work has been supported by Australian Research Council Discovery Project DP140103617.

\ifCLASSOPTIONcaptionsoff
  \newpage
\fi



%\bibliographystyle{IEEEtran}

\begin{thebibliography}{66}

\bibitem{Pearl2000}
J.~Pearl, \emph{Causality: models, reasoning, and inference}.\hskip 1em plus
  0.5em minus 0.4em\relax Cambridge University Press, 2000.


\bibitem{Spirtes2000}
P.~Spirtes, C.~Glymour, and R.~Scheines, \emph{{Causation, Prediction, and
  Search}}, 2nd~ed.\hskip 1em plus 0.5em minus 0.4em\relax Cambridge, MA: MIT
  Press, 2000.

\bibitem{pearl1988}
J.~Pearl, \emph{Probabilistic reasoning in intelligent systems: networks of
  plausible inference}.\hskip 1em plus 0.5em minus 0.4em\relax Morgan Kaufmann, 1988.

\bibitem{Neapolitan2004}
R.~Neapolitan, \emph{{Learning Bayesian networks}}.\hskip 1em plus 0.5em minus
  0.4em\relax Pearson Prenctice Hall, 2004.

\bibitem{granger1969}
C.~W. Granger, ``Investigating causal relations by econometric models and
  cross-spectral methods,'' \emph{Econometrica: Journal of the Econometric
  Society}, 424--438, 1969.

\bibitem{sims1972}
C.~A. Sims, ``Money, income, and causality,'' \emph{The American Economic
  Review}, vol.~62, no.~4, 540--552, 1972.

%\bibitem{Pearl2000causality}
%J.~Pearl, \emph{Causality: models, reasoning and inference}.\hskip 1em plus
%  0.5em minus 0.4em\relax Cambridge Univ Press, 2000, vol.~29.

\bibitem{Robinson1971}
J.~Robinson, ``{Counting labeled acyclic digraphs},'' in \emph{New Directions
  in the Theory of Graphs: Proc. of the Third Ann Arbor Conf. on Graph
  Theory}.\hskip 1em plus 0.5em minus 0.4em\relax New York: Academic Press,
  239--273, 1971.

\bibitem{Verma1990}
Verma and J.~Pearl, ``{Equivalence and synthesis of causal models},'' in
  \emph{Proceedings of the Sixth Conference on Uncertainty in Artificial
  Intelligence}, 220--227, 1990.



\bibitem{pcalg}
%\BIBentryALTinterwordspacing
M.~Kalisch, M.~M\"achler, D.~Colombo, M.~H. Maathuis, and P.~B\"uhlmann,
  ``Causal inference using graphical models with the {R} package {pcalg},''
  \emph{J STAT SOFW}, vol.~47, no.~11, 1--26, 2012.
  %[Online]. Available: \url{http://www.jstatsoft.org/v47/i11/}
%\BIBentrySTDinterwordspacing

\bibitem{bnlearn}
%\BIBentryALTinterwordspacing
M.~Scutari, ``Learning bayesian networks with the {bnlearn} {R} package,''
  \emph{J STAT SOFW}, vol.~35, no.~3, 1--22, 2010.
 % [Online]. Available: \url{http://www.jstatsoft.org/v35/i03/}
%\BIBentrySTDinterwordspacing

\bibitem{ZhangX2012}
X.~Zhang, X.~M.Liu, K.~He, et al. ``{Inferring gene regulatory networks from gene expression data by path consistency algorithm based on conditional mutual information,}" \emph{Bioinformatics}, vol.~28, no.~1, 98-104, 2012.


\bibitem{Maathuis2010}
H.~M Maathuis, D.~Colombo, M.~Kalisch, P.~Buhlmann, ``{Predicting causal effects
  in large-scale systems from observational data}," \emph{Nature Methods}, 7: 247--249, 2010.

  \bibitem{le2013}
T.~D. Le, L.~Liu, A.~Tsykin, G.~J. Goodall, B.~Liu, B.-Y. Sun, and J.~Li,
  ``{Inferring microRNA--mRNA causal regulatory relationships from expression
  data},'' \emph{Bioinformatics}, vol.~29, no.~6, 765--771, 2013.

 \bibitem{Zhang2014a}
J.~Zhang, T.~D. Le, L.~Liu, B.~Liu, J.~He, \emph{et~al.}, ``{Inferring condition-specific
  miRNA activity from matched miRNA and mRNA expression data},"
\emph{Bioinformatics}, 30: 3070-7, 2014.
%\bibAnnoteFile{Zhang2014a}

\bibitem{Zhang2014b}
J.~Zhang, T.~D. Le, L.~Liu, B.~Liu, J.~He, \emph{et~al.}, ``{Identifying direct
  miRNA-mRNA causal regulatory relationships in heterogeneous data},"
\emph{Journal of Biomedical Informatics}, 52: 438-47, 2014.
%\bibAnnoteFile{Zhang2014b}


\bibitem{silverstein2000}
C.~Silverstein, S.~Brin, R.~Motwani, and J.~Ullman, ``Scalable techniques for
  mining causal structures,'' \emph{Data Min. Knowl. Discov.},
  vol.~4, no.~2, 163--192, 2000.

\bibitem{tsamardinos2003}
I.~Tsamardinos, C.~F. Aliferis, A.~Statnikov, and E.~Statnikov, ``Algorithms
  for large scale markov blanket discovery,'' in \emph{Proceedings of the
  sixteenth international Florida artificial intelligence research society
  conference}, 376--381, 2003.

\bibitem{li2013}
J.~Li, T.~D. Le, L.~Liu, J.~Liu, Z.~Jin, and B.~Sun, ``Mining causal
  association rules,'' in \emph{Proceedings of ICDM Workshop on Causal
  Discovery (CD)}, 2013.

  \bibitem{li2015Tist}
J.~Li, T.~D. Le, L.~Liu, J.~Liu, Z.~Jin, B.~Sun, and S.~Ma, ``From Observational Studies to Causal Rule Mining,'' in \emph{ACM Transactions on Intelligent Systems and Technology (TIST) }, vol.~7, no.~3, 2015.

\bibitem{colombo2012}
%\BIBentryALTinterwordspacing
D.~{Colombo} and M.~H. {Maathuis}, ``{A modification of the PC algorithm
  yielding order-independent skeletons},'' \emph{ArXiv e-prints}, Nov. 2012.
  %provided by the SAO/NASA Astrophysics Data System.
  % [Online]. Available: \url{{http://adsabs.harvard.edu/abs/2012arXiv1211.3295C}}
%\BIBentrySTDinterwordspacing


\bibitem{dash1999hybrid}
D.~Dash and M.~J. Druzdzel, ``A hybrid anytime algorithm for the construction
  of causal models from sparse data,'' in \emph{Proceedings of the Fifteenth
  conference on Uncertainty in artificial intelligence}.\hskip 1em plus 0.5em
  minus 0.4em\relax Morgan Kaufmann Publishers Inc., 142--149, 1999.

\bibitem{cano2008score}
A.~Cano, M.~G{\'o}mez-Olmedo, and S.~Moral, ``A score based ranking of the
  edges for the pc algorithm,'' in \emph{Proceedings of the Fourth European
  Workshop on Probabilistic Graphical Models}, 41--48, 2008.

\bibitem{dean2008mapreduce}
J.~Dean and S.~Ghemawat, ``Mapreduce: simplified data processing on large
  clusters,'' \emph{Communications of the ACM}, vol.~51, no.~1, 107--113,
  2008.


\bibitem{Heckerman1995}
D.~Heckerman, D.~Geiger, and D.~M. Chickering, ``Learning bayesian networks:
  The combination of knowledge and statistical data,'' \emph{Machine learning},
  vol.~20, no.~3, 197--243, 1995.

\bibitem{edwards2000}
D.~Edwards, \emph{Introduction to graphical modelling}.\hskip 1em plus 0.5em
  minus 0.4em\relax Springer, 2000.

\bibitem{cooper1992}
G.~F. Cooper and E.~Herskovits, ``A bayesian method for the induction of
  probabilistic networks from data,'' \emph{Machine learning}, vol.~9, no.~4,
  pp. 309--347, 1992.

\bibitem{glymour1999}
C.~Glymour and G.~Cooper, ``Causation, computation and discovery,'' 1999.

\bibitem{chickering1994}
D.~M. Chickering, D.~Geiger, D.~Heckerman \emph{et~al.}, ``Learning bayesian
  networks is np-hard,'' Citeseer, Tech. Rep., 1994.

\bibitem{cooper1997}
G.~F. Cooper, ``A simple constraint-based algorithm for efficiently mining
  observational databases for causal relationships,'' \emph{Data Min. Knowl. Discov.}, vol.~1, no.~2, 203--224, 1997.

\bibitem{Kalisch2007}
M.~Kalisch and P.~Buhlmann, ``{Estimating high-dimensional directed acyclic
  graphs with the PC-algorithm},'' \emph{JMLR},
  vol.~8, 613--636, 2007.

\bibitem{steck1999bayesian}
H.~Steck and V.~Tresp, ``Bayesian belief networks for data mining,'' in
  \emph{Proceedings of the 2. Workshop on Data Mining und Data Warehousing als
  Grundlage moderner entscheidungsunterst{\"u}tzender Systeme}.\hskip 1em plus
  0.5em minus 0.4em\relax Citeseer, 145--154, 1999.

\bibitem{Abellánsomevariations}
J.~Abellán, M.~Gómez-olmedo, and S.~Moral, ``Some variations on the pc
  algorithm,'' in \emph{Proceedings of the first European Workshop on
  Probabilistic Graphical Models}, 2006.

\bibitem{Margaritis99a}
D.~Margaritis and S.~Thrun, ``{B}ayesian network induction via local
  neighborhoods,'' in \emph{Proceedings of Conference on Neural Information
  Processing Systems (NIPS-12)}, S.~Solla, T.~Leen, and K.~R. M\"{u}ller,
  Eds.\hskip 1em plus 0.5em minus 0.4em\relax MIT Press, 1999.

\bibitem{yaramakala2005}
S.~Yaramakala and D.~Margaritis, ``Speculative markov blanket discovery for
  optimal feature selection,'' in \emph{Data Mining, Fifth IEEE International
  Conference on}.\hskip 1em plus 0.5em minus 0.4em\relax IEEE, 2005.

\bibitem{aliferis2010a}
C.~F. Aliferis, A.~Statnikov, I.~Tsamardinos, S.~Mani, and X.~D. Koutsoukos,
  ``Local causal and markov blanket induction for causal discovery and feature
  selection for classification part i: Algorithms and empirical evaluation,''
  \emph{JMLR}, vol.~11, 171--234, 2010.

\bibitem{Tsamardinos2003MMMB}
%\BIBentryALTinterwordspacing
I.~Tsamardinos, C.~F. Aliferis, and A.~Statnikov, ``Time and sample efficient
 discovery of markov blankets and direct causal relations,'' in
  \emph{Proceedings of the ninth ACM SIGKDD international conference on
  Knowledge discovery and data mining}, ser. KDD '03.\hskip 1em plus 0.5em
  minus 0.4em\relax New York, NY, USA: ACM, 673--678, 2003.
  % [Online]. Available: \url{http://doi.acm.org/10.1145/956750.956838}
%\BIBentrySTDinterwordspacing
%

\bibitem{pena2007towards}
J.~M. Pe{\~n}a, R.~Nilsson, J.~Bj{\"o}rkegren, and J.~Tegn{\'e}r, ``Towards
  scalable and data efficient learning of markov boundaries,''
  \emph{INT J APPROX REASON}, vol.~45, no.~2, 211--232, 2007.

\bibitem{fu2008}
S.~Fu and M.~C. Desmarais, ``Fast markov blanket discovery algorithm via local
  learning within single pass,'' in \emph{Advances in Artificial
  Intelligence}.\hskip 1em plus 0.5em minus 0.4em\relax Springer, 96--107, 2008.

\bibitem{jin2012}
Z.~Jin, J.~Li, L.~Liu, T.~D. Le, B.~Sun, and R.~Wang, ``Discovery of causal
  rules using partial association,'' in \emph{Data Mining (ICDM), 2012 IEEE
  12th International Conference on}.\hskip 1em plus 0.5em minus 0.4em\relax
  IEEE, 2012, 309--318.

\bibitem{li2015}
J.~Li, L.~Liu, T.~D. Le, \emph{Practical approaches to causal relationship exploration}. Springer, 2015.

\bibitem{Scutari}
M.~Scutari, ``{Bayesian Network Constraint-Based Structure Learning Algorithms: Parallel and Optimised Implementations in the bnlearn R Package}," \emph{arXiv 1406:7648}, 2014.

\bibitem{Chen}
Y.~Chen, J.~Tian, O.~Nikolova, S.~Aluru, ``{A Parallel Algorithm for Exact Bayesian Structure Discovery in Bayesian Networks}," \emph{ http://arxiv.org/abs/1408.1664v1}, 2014.

\bibitem{ramsey2012adjacency}
J.~Ramsey, J.~Zhang, and P.~L. Spirtes, ``Adjacency-faithfulness and
  conservative causal inference,'' \emph{arXiv preprint arXiv:1206.6843}, 2012.

\bibitem{R}
%\BIBentryALTinterwordspacing
{R Core Team}, \emph{R: A Language and Environment for Statistical Computing},
  R Foundation for Statistical Computing, Vienna, Austria, 2014.
 % [Online]. Available: \url{http://www.R-project.org/}
%\BIBentrySTDinterwordspacing

%\bibitem{monks1991}
%A.~Monks, D.~Scudiero, P.~Skehan, R.~Shoemaker, K.~Paull, D.~Vistica, C.~Hose,
%  J.~Langley, P.~Cronise, A.~Vaigro-Wolff \emph{et~al.}, ``Feasibility of a
%  high-flux anticancer drug screen using a diverse panel of cultured human
%  tumor cell lines,'' \emph{Journal of the National Cancer Institute}, vol.~83,
%  no.~11, pp. 757--766, 1991.

%\bibitem{kilde2011}
%R.~S{\o}kilde, B.~Kaczkowski, and A.~Podolska, ``{Global microRNA analysis of
%  the NCI-60 cancer cell panel},'' \emph{Mol. Can. Ther.},
%  vol.~10, pp. 375--384, 2011.



\bibitem{Le2014}
T.~D Le, L. Liu, J.~Zhang, B.~Liu, J.~Li, ``{From miRNA regulation to miRNA - TF
  co-regulation: computational approaches and challenges}," \emph{Briefings in Bioinformatics},  16: 475-96, 2015.

%\bibAnnoteFile{Maathuis2010}
%\bibitem{lu2005}
%J.~Lu, G.~Getz, E.~A. Miska, E.~Alvarez-Saavedra, J.~Lamb, D.~Peck,
%  A.~Sweet-Cordero, B.~L. Ebert, R.~H. Mak, A.~A. Ferrando \emph{et~al.},
 % ``{MicroRNA expression profiles classify human cancers},'' \emph{Nature},
 % vol. 435, no. 7043, pp. 834--838, 2005.

%\bibitem{ramaswamy2001}
%S.~Ramaswamy, P.~Tamayo, R.~Rifkin, S.~Mukherjee, C.-H. Yeang, M.~Angelo,
%  C.~Ladd, M.~Reich, E.~Latulippe, J.~P. Mesirov \emph{et~al.}, ``Multiclass
%  cancer diagnosis using tumor gene expression signatures,'' \emph{Proceedings
%  of the National Academy of Sciences}, vol.~98, no.~26, pp. 15\,149--15\,154,
%  2001.

%\bibitem{riaz2013}
%M.~Riaz, M.~T. van Jaarsveld, A.~Hollestelle, W.~J. Prager-van~der Smissen,
%  A.~A. Heine, A.~W. Boersma, J.~Liu, J.~Helmijr, B.~Ozturk, M.~Smid
%  \emph{et~al.}, ``{miRNA expression profiling of 51 human breast cancer cell
 % lines reveals subtype and driver mutation-specific miRNAs},'' \emph{Breast
 % Cancer Research}, vol.~15, no.~2, p. R33, 2013.

\bibitem{marbach2012}
D.~Marbach, J.~C. Costello, R.~K{\"u}ffner, N.~M. Vega, R.~J. Prill, D.~M.
  Camacho, K.~R. Allison, M.~Kellis, J.~J. Collins, G.~Stolovitzky
  \emph{et~al.}, ``{Wisdom of crowds for robust gene network inference},''
  \emph{Nature Methods}, vol.~9, 796--804, 2012.



%%%%%%%%%%%%%%%%%%%%%%%%%%
\bibitem{Bartel2004}
D.~P Bartel ``{MicroRNAs: genomics, biogenesis, mechanism, and function},".
\emph{Cell}, 116: 281--297, 2004.

\bibitem{KimVN2006}
V.~N Kim, J.~W Nam ``{Genomics of microRNA},".
\emph{Trends Genetics}, 22: 165-173, 2006.

\bibitem{enright2004}
A.~J Enright, B.~John, U.~Gaul, T.~Tuschl, C.~Sander C, \emph{et~al.}, ``{microRNA targets
  in Drosophila},".
\emph{Genome Biology}, 5: R1--R1, 2004.

\bibitem{Lewis2005}
B.~P Lewis, C.~B Burge, D.~P Bartel, ``{Conserved seed pairing, often flanked by
  adenosines, indicates that thousands of human genes are microRNA targets},"
\emph{Cell}, 120: 15--20, 2005.
%\bibAnnoteFile{Lewis2005}

\bibitem{Krek2005}
A.~Krek, D.~Gr$\ddot{u}$n, M.~N Poy, R.~Wolf, L.~Rosenberg, \emph{et~al.},  ``{Combinatorial microRNA target predictions},"
\emph{Nature Genetics}, 37: 495-500, 2005.
%\bibAnnoteFile{Krek2005}

\bibitem{Rajewsky2006}
N.~Rajewsky, ``{microRNA target predictions in animals},"
\emph{Nature Genetics}, 38: S8-13, 2006.
%\bibAnnoteFile{Rajewsky2006}
\bibitem{liuH2010}
H.~Liu, A.~Brannon, A.~Reddy, G.~Alexe, M.~Seiler, \emph{et~al.}, ``{Identifying mRNA
  targets of microRNA dysregulated in cancer: with application to clear cell
  Renal Cell Carcinoma},"
\emph{BMC Systems Biology}, 4: 51, 2010.
%\bibAnnoteFile{liuH2010}

\bibitem{van2010}
I.~Van~der Auwera, R.~Limame, P.~Van~Dam, P.~Vermeulen, L.~Dirix, \emph{et~al.}, ``{Integrated miRNA and mRNA expression profiling of the inflammatory breast
  cancer subtype},"
\emph{British Journal of Cancer}, 103: 532--541, 2010.
%\bibAnnoteFile{van2010}

\bibitem{lu2011}
Y.~Lu, Y.~Zhou, W.~Qu, M.~Deng, C.~Zhang, ``{A Lasso regression model for the
  construction of microRNA-target regulatory networks},"
\emph{Bioinformatics}, 27: 2406--2413, 2011.
%\bibAnnoteFile{lu2011}

\bibitem{muniategui2012}
A.~Muniategui, R.~Nogales-Cadenas, M.~V{\'a}zquez, X.~L Aranguren, X.~Agirre, \emph{et~al.},
   ``{Quantification of miRNA-mRNA interactions},"
\emph{PloS One}, 7: e30766, 2012.
%\bibAnnoteFile{muniategui2012}

\bibitem{Joung2007}
J.~G Joung, K.~B Hwang, J.W Nam, S.J Kim, B.~T Zhang ``{Discovery of microRNA-mRNA
  modules via population-based probabilistic learning},"
\emph{Bioinformatics}, 23: 1141--7, 2007.
%\bibAnnoteFile{Joung2007}

\bibitem{Tran2008}
D.~H Tran, K.~Satou, T.~B Ho ``{Finding microRNA regulatory modules in human
  genome using rule induction.},"
\emph{BMC Bioinformatics}, 9 Suppl 12: S5, 2008.
%\bibAnnoteFile{Tran2008}

\bibitem{Huang2007}
J.~C Huang, T.~Babak, T.~W Corson, G.~Chua, S.~Khan, \emph{et~al.}, ``{Using expression
  profiling data to identify human microRNA targets},"
\emph{Nature Methods}, 4: 1045--1050, 2007.
%\bibAnnoteFile{Huang2007}

\bibitem{liu2009a}
B.~Liu, J.~Li, A.~Tsykin, L.~Liu, A.~B Gaur, \emph{et~al.} ``{Exploring complex
  miRNA-mRNA regulatory networks by splitting-averaging strategy},"
\emph {BMC Bioinformatics}, 19: 1--19, 2009.
%\bibAnnoteFile{liu2009a}


\bibitem{le2013b}
T.~D. Le, L.~Liu, B.~Liu, A.~Tsykin, G.~J. Goodall, K.~Satou, J.~Li, ``{Inferring microRNA
  and transcription factor regulatory networks in heterogeneous data,}" \emph{BMC Bioinformatics}, vol.~14:92, 2013.



 \bibitem{Le2015}
T.~D. Le, J.~Zhang, L.~Liu, J.~Li, ``{Ensemble Methods for MiRNA Target Prediction from Expression Data},"
\emph{PloS one}, 10(6), e0131627, 2015.

\bibitem{Maathuis2009}
H.~M. Maathuis, M.~Kalisch, and P.~Buhlmann, ``{Estimating high-dimensional
  intervention effects from observational data},'' \emph{Annals of Statistics},
  vol.~37, no.~6, 3133--3164, 2009.


\bibitem{vergoulis2012}
T.~Vergoulis, I.~S. Vlachos, P.~Alexiou, G.~Georgakilas, M.~Maragkakis, \emph{et~al.}, ``{Tarbase 6.0: capturing the exponential growth of miRNA targets with
  experimental support},"
\emph{Nucleic Acids Research}, 40: D222--D229, 2012.
%\bibAnnoteFile{vergoulis2012}

\bibitem{xiao2009}
F.~Xiao, Z.~Zuo, G.~Cai, S.~Kang, X.~Gao, \emph{et~al.}, ``{mirecords: an integrated
  resource for microrna-target interactions},"
\emph{Nucleic Acids Research}, 37: D105--D110, 2009.
%\bibAnnoteFile{xiao2009}

\bibitem{dweep2011}
H.~Dweep, C.~Sticht, P.~Pandey, N.~Gretz, ``{mirwalk--database: prediction of
  possible mirna binding sites by walking the genes of three genomes},"
\emph{Journal of Biomedical Informatics}, 44: 839--847, 2011.
%\bibAnnoteFile{dweep2011}

\bibitem{HsuSD2014}
S.~D Hsu, Y.~T Tseng, S.~Shrestha, Y.~L Lin, A.~Khaleel, \emph{et~al.},  ``{miRTarBase
  update 2014: an information resource for experimentally validated
  miRNA-target interactions},"
\emph{Nucleic Acids Research}, 42: D78-85, 2014.
%\bibAnnoteFile{HsuSD2014}
%%%%%%%%%%%%%%%%%%%%%%%%%%%%%%%


\bibitem{colombo2012learning}
D.~Colombo, M.~H. Maathuis, M.~Kalisch, T.~S. Richardson \emph{et~al.},
  ``Learning high-dimensional directed acyclic graphs with latent and selection
  variables,'' \emph{Annals of Statistics}, vol.~40, no.~1, 294--321,
  2012.

\bibitem{richardson1996discovery}
T.~Richardson, ``A discovery algorithm for directed cyclic graphs,'' in
  \emph{Proceedings of the Twelfth international conference on Uncertainty in
  artificial intelligence}.\hskip 1em plus 0.5em minus 0.4em\relax Morgan
  Kaufmann Publishers Inc., 1996, 454--461.



\end{thebibliography}


\begin{IEEEbiography}[{\includegraphics[width=1in,height=1.25in,clip,keepaspectratio]{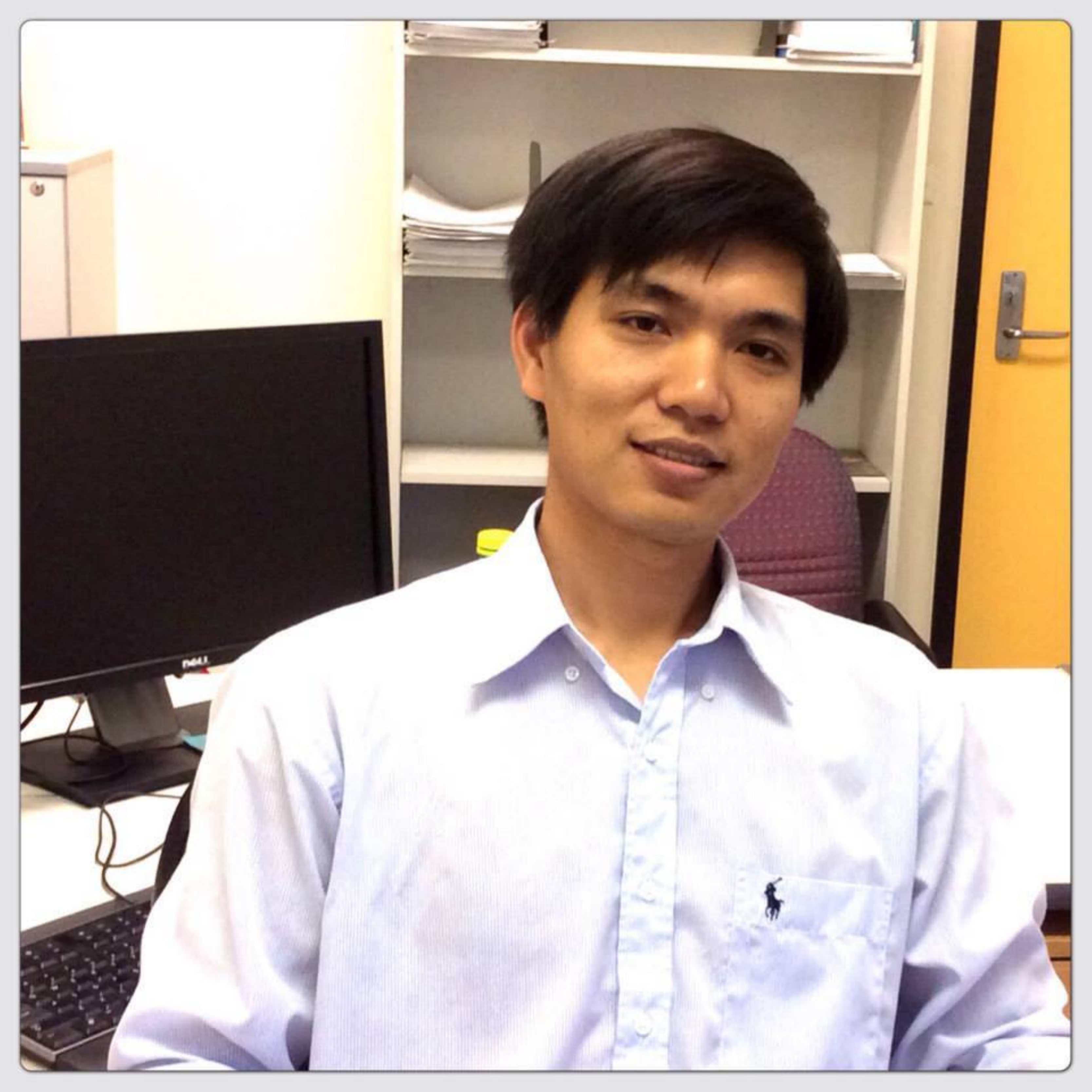}}]{Thuc Duy Le}
%\begin{IEEEbiography}{Thuc Duy Le}
is a research fellow at the University of South Australia (UniSA). He received his BSc (2002) and MSc (2006) in pure Mathematics from the University of Pedagogy, Ho Chi Minh City, Vietnam, and BSc (2010) in Computer Science from UniSA. He received his PhD in Computer Science (Bioinformatics) from UniSA in 2014. His research interests are Bioinformatics, data mining, and machine learning.
\end{IEEEbiography}

\begin{IEEEbiography}[{\includegraphics[width=1in,height=1.25in,clip,keepaspectratio]{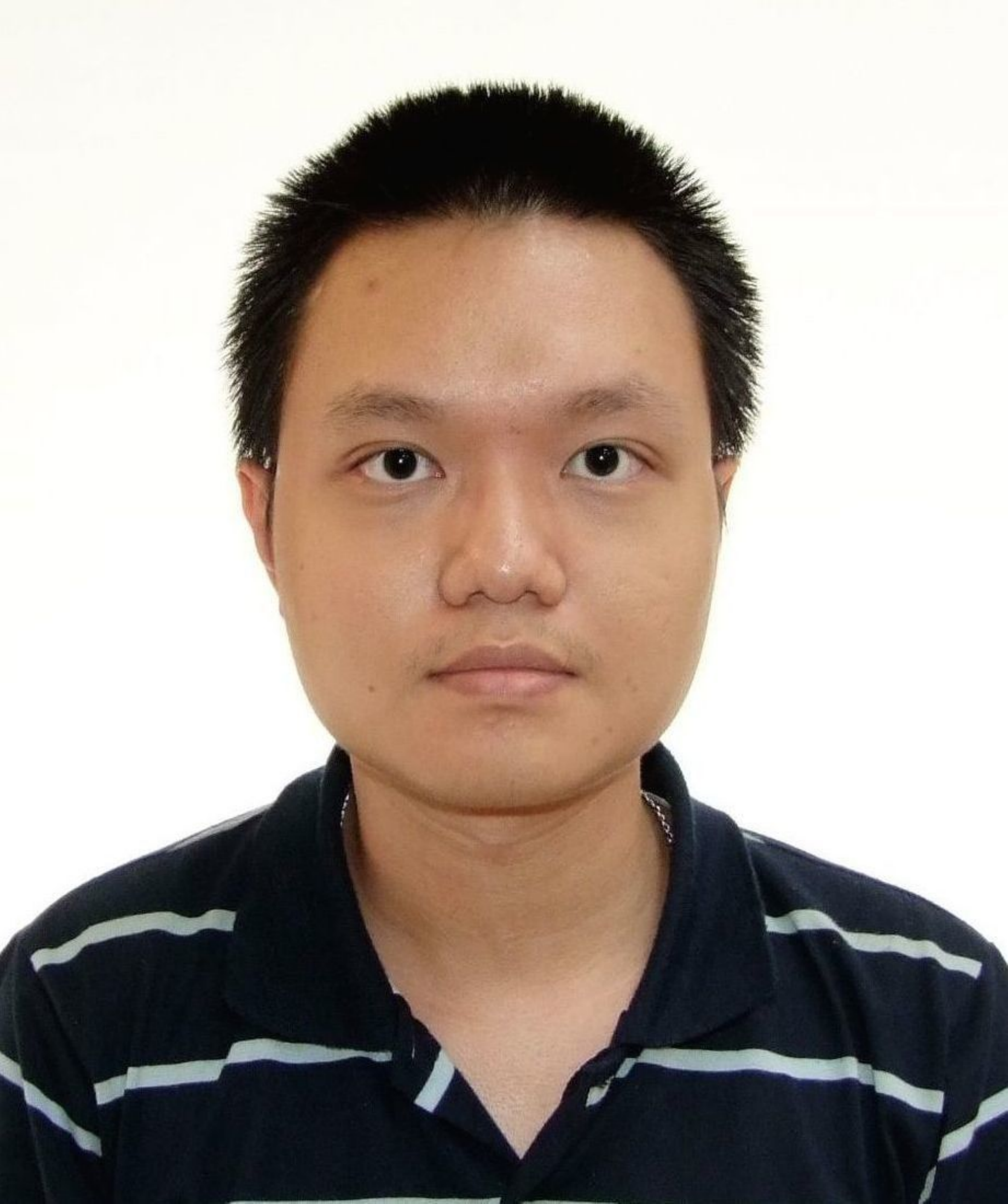}}]{Tao Hoang}
%\begin{IEEEbiography}{Thuc Duy Le}
is a PhD student at the University of South Australia (UniSA). He received his BSc (2012) in Computer Science from the National University of Singapore (NUS), Singapore. His research interests are health informatics, text mining and machine learning.
\end{IEEEbiography}

\begin{IEEEbiography}[{\includegraphics[width=1in,height=1.25in,clip,keepaspectratio]{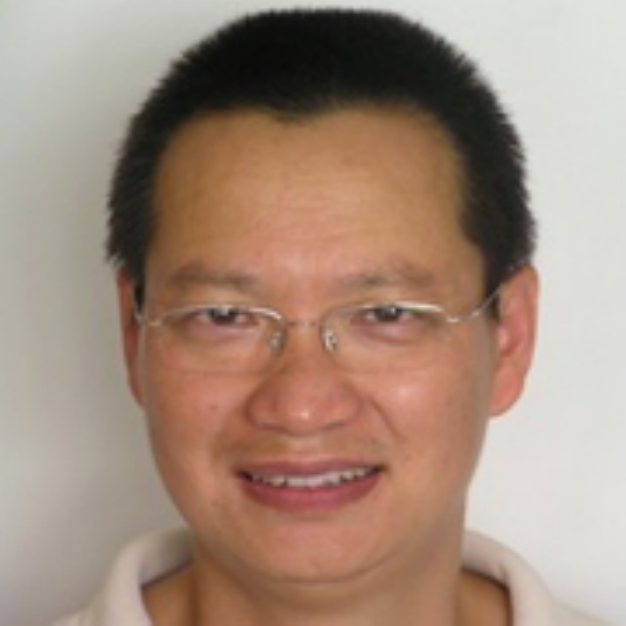}}]{Jiuyong Li}
%\begin{IEEEbiography}{Thuc Duy Le}
Jiuyong Li is a professor at the School of Information Technology and Mathematical Sciences, University of South Australia. He received his PhD degree in computer science from the Griffith University, Australia. His research interests are in the field of data mining, privacy preserving and Bioinformatics. He has published more than 100 papers and his research has been supported by 5 Australian Research Council Discovery grants.
\end{IEEEbiography}
%\begin{IEEEbiographynophoto}{Tao Hoang}
%is a PhD student at the University of South Australia. He graduated from the National University of Singapore in 2012 and worked as a research software engineering in AStar from 2012-2014.

%\newpage

\begin{IEEEbiography}[{\includegraphics[width=1in,height=1.25in,clip,keepaspectratio]{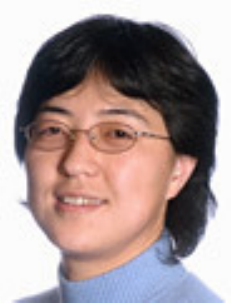}}]{Lin Liu}
is a senior lecturer at the School of Information Technology and Mathematical Sciences, University of South Australia (UniSA). She received her bachelor and master degrees in Electronic Engineering from Xidian University, China in 1991 and 1994 respectively, and her PhD degree in computer systems engineering from UniSA in 2006. Dr Liu’s research interests include data mining and bioinformatics, as well as Petri nets and their applications to protocol verification and network security analysis.
\end{IEEEbiography}

\begin{IEEEbiography}[{\includegraphics[width=1in,height=1.25in,clip,keepaspectratio]{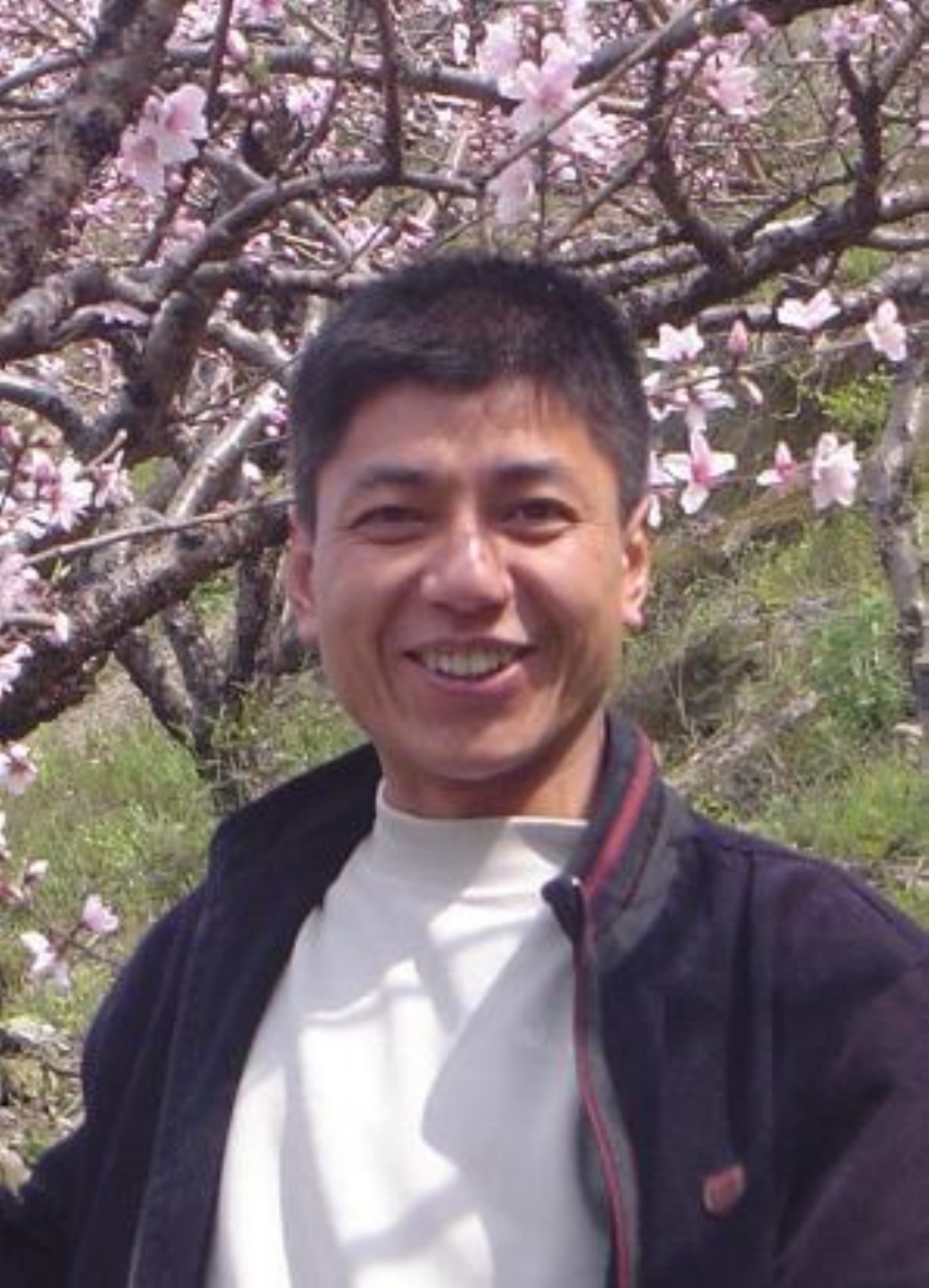}}]{Huawen Liu}
%\begin{IEEEbiography}{Thuc Duy Le}
is an associate professor at the School of Mathematics, Physics and Information Engineering, Zhejiang Normal University, China. He received his MSc and PhD degree in computer science from Jilin University, China in 2007 and 2010 respectively. Since 2010, he is with the Department of Computer Science at Zhejiang Normal University, China. His interests include data mining, feature selection and sparse learning for machine learning
\end{IEEEbiography}

\begin{IEEEbiography}[{\includegraphics[width=1in,height=1.25in,clip,keepaspectratio]{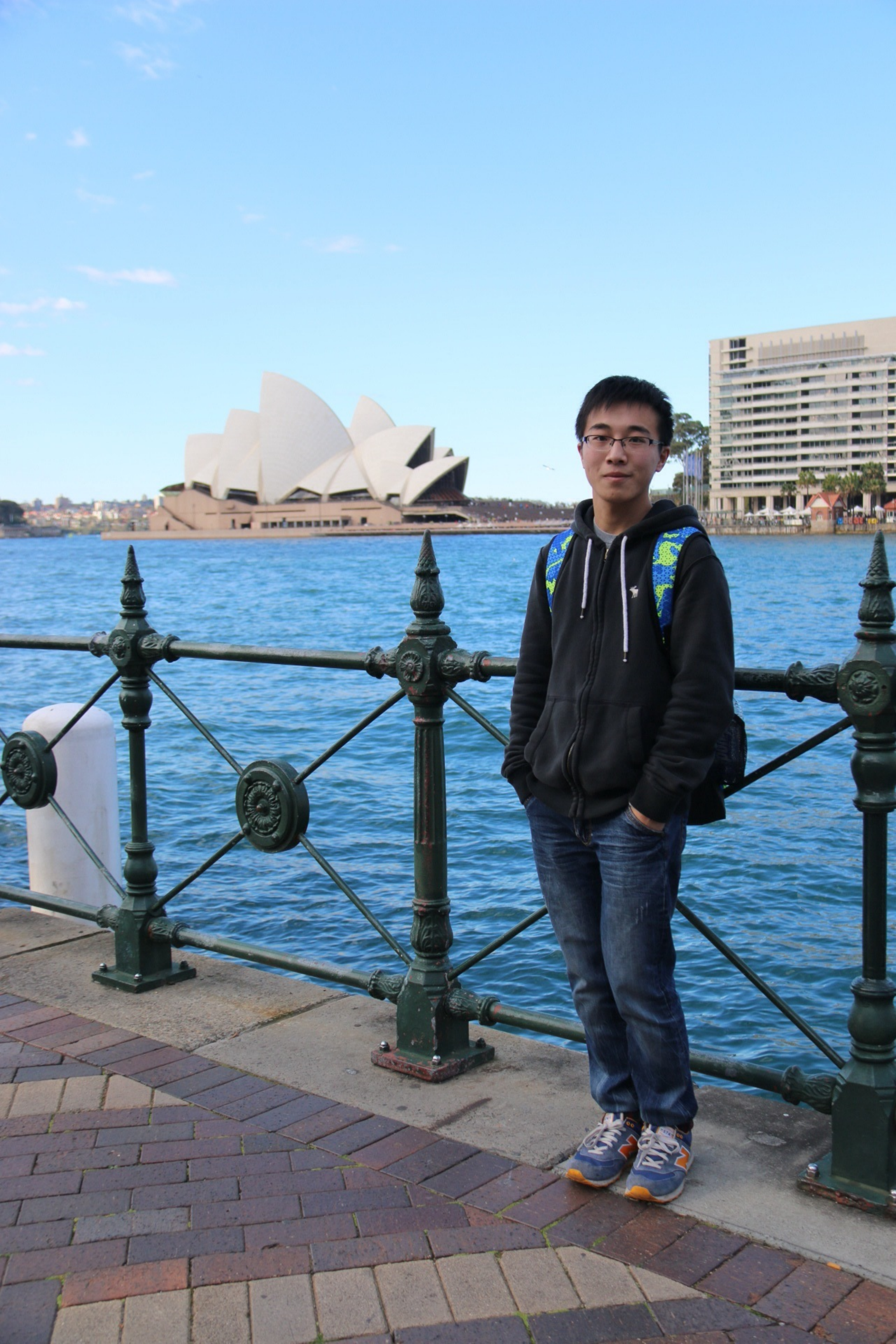}}]{Shu Hu}
%\begin{IEEEbiography}{Thuc Duy Le}
is a visiting researcher at the University of South Australia (UniSA). He received his BSc(2013) in Electronics and Information Engineering from North China Institute of Science and Technology. His research interests are data mining and machine learning.
\end{IEEEbiography}

%\vfill


\end{document}